%% file: main.tex
\documentclass[10pt,twocolumn,letterpaper]{article}

\usepackage{cvpr}              
\usepackage{svg}
\usepackage{multirow}
\usepackage{graphicx}
\definecolor{cvprblue}{rgb}{0.21,0.49,0.74}
\usepackage[pagebackref,breaklinks,colorlinks,citecolor=cvprblue]{hyperref}

\def\confName{CVPR}
\def\confYear{2024}
\begin{document}

\title{MetaFood CVPR 2024 Challenge on Physically Informed 3D Food Reconstruction: Methods and Results}

\author{%
\textbf{Workshop and Challenge Organizers}\\\vspace{0.5em}
Jiangpeng He\textsuperscript{1},
Yuhao Chen\textsuperscript{2}, 
Gautham Vinod\textsuperscript{1},
Talha Ibn Mahmud\textsuperscript{1}\\
Fengqing Zhu\textsuperscript{1}, 
Edward Delp\textsuperscript{1},
Alexander Wong\textsuperscript{2}, 
Pengcheng Xi\textsuperscript{3}\\[0.5em]
\textsuperscript{1} Purdue University
\textsuperscript{2} University of Waterloo
\textsuperscript{3} National Research Council Canada \\[1em]
\textbf{Challenge Winner Team - First Place}\\\vspace{0.5em}
Ahmad AlMughrabi\textsuperscript{4}, 
Umair Haroon\textsuperscript{4}, 
Ricardo Marques\textsuperscript{4}, 
Petia Radeva\textsuperscript{4}\\[0.5em]
\textsuperscript{4} Universitat de Barcelona \\[1em]
\textbf{Challenge Winner Team - Second Place}\\\vspace{0.5em}
Jiadong Tang\textsuperscript{5}, 
Dianyi Yang\textsuperscript{5}, 
Yu Gao\textsuperscript{5}, 
Zhaoxiang Liang\textsuperscript{5}\\[0.5em]
\textsuperscript{5} Beijing Institute of Technology \\[1em]
\textbf{Challenge Winner Team - Best 3D Mesh Reconstruction}\\\vspace{0.5em}
Yawei Jueluo\textsuperscript{6}, 
Chengyu Shi\textsuperscript{7}, 
Pengyu Wang\textsuperscript{8}\\[0.5em]
\textsuperscript{6} Baidu Inc.
\textsuperscript{7} XPeng Motors
\textsuperscript{8} Beijing University of Posts and Telecommunications
}

\maketitle

\input{sec/0_abstract}

\input{sec/1_intro}

\input{sec/2_relatedwork}
\input{sec/2_Dataset}
\input{sec/VolETA}
\input{sec/ININ-VIAUN}
\input{sec/FoodRiddle}
\input{sec/5_conclusion}
{
    \small
    \bibliographystyle{ieeenat_fullname}
    \bibliography{main}
}
\end{document}

%% file: sec/0_abstract.tex
\begin{abstract}
The increasing interest in computer vision applications for nutrition and dietary monitoring has led to the development of advanced 3D reconstruction techniques for food items. However, the scarcity of high-quality data and limited collaboration between industry and academia have constrained progress in this field. Building on recent advancements in 3D reconstruction, we host the MetaFood Workshop and its challenge for Physically Informed 3D Food Reconstruction. This challenge focuses on reconstructing volume-accurate 3D models of food items from 2D images, using a visible checkerboard as a size reference. Participants were tasked with reconstructing 3D models for 20 selected food items of varying difficulty levels: easy, medium, and hard. The easy level provides 200 images, the medium level provides 30 images, and the hard level provides only 1 image for reconstruction. In total, 16 teams submitted results in the final testing phase. The solutions developed in this challenge achieved promising results in 3D food reconstruction, with significant potential for improving portion estimation for dietary assessment and nutritional monitoring. More details about this workshop challenge and access to the dataset can be found at \url{https://sites.google.com/view/cvpr-metafood-2024}.
\end{abstract}

%% file: sec/1_intro.tex
\section{Introduction}
\label{sec:intro}
The intersection of computer vision and culinary arts has opened new frontiers in dietary monitoring and nutritional analysis. The CVPR 2024 MetaFood Workshop Challenge represents a significant step in this direction, addressing the growing need for accurate, scalable methods of food portion estimation and nutritional intake tracking. These technologies are crucial to promote healthy eating habits and managing diet-related health conditions. 

By focusing on reconstructing accurate 3D models of food items from both multi-view and single-view inputs, this challenge aims to bridge the gap between existing methods and real-world requirements. The challenge encourages the development of innovative techniques that can handle the complexities of food shapes, textures, and lighting conditions, while also addressing the practical constraints of real-world dietary assessment scenarios. By bringing together researchers and practitioners in computer vision, machine learning, and nutrition science, this challenge seeks to catalyze advancements in 3D food reconstruction that could significantly improve the accuracy and applicability of food portion estimation in various contexts, from personal health monitoring to large-scale nutritional studies. 

Traditional diet assessment methods~\cite{thompson2017dietary}, such as 24-Hour Recall or Food Frequency Questionnaire (FFQ), often rely on manual input, which can be inaccurate and cumbersome. Furthermore, the absence of 3D information in 2D RGB food images presents significant challenges for regression-based methods~\cite{9733557, he2020multitask} that estimate food portions directly from eating occasion images. By advancing 3D reconstruction techniques for food items, we aim to offer more precise and intuitive tools for nutritional assessment. This technology has the potential to improve the sharing of food experiences and significantly impact fields such as nutrition science and public health. 

\begin{figure}[t]
    \centering
    \includegraphics[width=\linewidth]{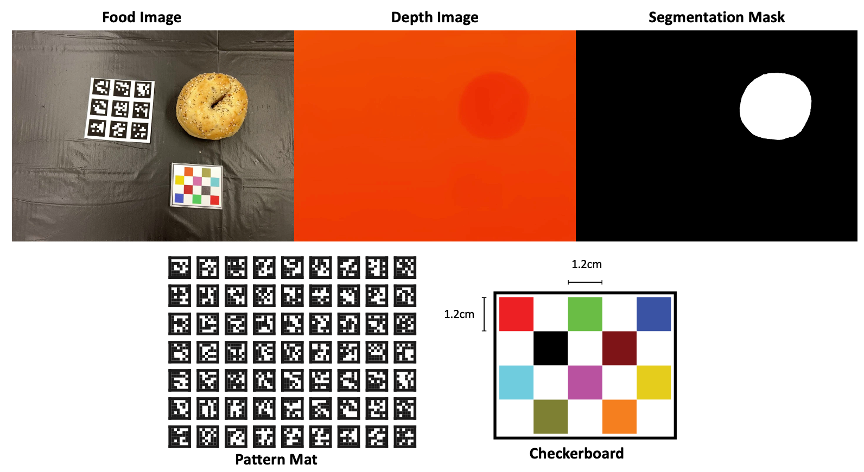}
    \caption{Sample challenge data for “everything bagel”.}
    \label{fig:1}
\end{figure}

The challenge tasked participants with reconstructing 3D models of 20 selected food items from 2D images, simulating a scenario where a cellphone with a depth camera is used for food logging and nutritional monitoring. It was structured into three difficulty levels. The easy level food object provided approximately 200 frames uniformly sampled from video, the medium level offered about 30 images, and the hard level presented participants with a single monocular top-view image. This structure was designed to test the robustness and versatility of the proposed solutions across various real-world scenarios. A sample hard food object is shown in Figure~\ref{fig:1}. The key features of the challenge include the use of a visible checkerboard as a physical reference, as well as the availability of the depth image for each video frame, ensuring that the reconstructed 3D models maintain accurate real-world scaling for portion size estimation.

This challenge not only pushes the boundaries of 3D reconstruction technology, but also paves the way for more accurate, robust, and user-friendly applications in the real world, such as image-based dietary assessment. The solutions developed here have the potential to significantly impact how we monitor and understand our nutritional intake, contributing to broader goals in health and wellness. As we continue to advance in this field, we anticipate seeing innovative applications that could revolutionize personal health management, nutritional research, and the food industry at large. The structure of this technical report is as follows. In Section~\ref{sec:related work}, we review existing related work for food portion size estimation. In Section~\ref{sec: dataset}, we introduce the dataset used for this challenge and the detailed evaluation pipelines. Finally, we summarize the methodology and experimental results for the three winner teams (\textit{VolETA}, \textit{ININ-VIAUN}, \textit{FoodRiddle}) in Section~\ref{sec:VolETA}, Section~\ref{sec: ININ-VIAUN}, and Section~\ref{sec: FoodRiddle}, respectively.

%% file: sec/2_relatedwork.tex
\section{Related Work}
\label{sec:related work}

Food portion estimation is an important component of image-based dietary assessment~\cite{shao2021_ibdasystem, he2021end} with the goal of estimating the volume, energy or macronutrients directly from the input eating occasion images. Compared to the widely studied food recognition task~\cite{min2023large, mao2021visual, mao2021_nutri_hierarchy, He_2021_ICCVW, pan2023personalized, he2022long, huang2024automatic}, food portion estimation presents a unique challenge due to the absence of 3D information and physical references, which are essential for accurately inferring the real-world size of food portions. Specifically, accurately estimating portion sizes requires an understanding of the volume and density of the food, aspects that cannot be easily determined from a two-dimensional image, which highlights the need for advanced methodologies and technologies to address this issue. Existing food portion estimation methods can be categorized into four main groups \cite{Lo2020}.

\noindent\textbf{Stereo-Based Approach.} These methods rely on multiple frames to reconstruct the 3D structure of the food. In \cite{5403087}, food volume is estimated using multi-view stereo reconstruction based on epipolar geometry. Similarly, \cite{6395199} performs two-view dense reconstruction. Simultaneous Localization and Mapping (SLAM) is utilized in \cite{8329671} for continuous and real-time food volume estimation. The primary limitation of these methods is the requirement for multiple images, which is impractical for real-world deployment.

\noindent\textbf{Model-Based Approach.} Predefined shapes and templates are leveraged to estimate the target volume. For instance, \cite{6738522} assigns certain templates to foods from a library and applies transformations based on physical references to estimate the size and location of the food. A similar template matching approach is used in \cite{jia20123d} to estimate food volume from a single image. However, these methods cannot accommodate variations in food shapes that deviate from predefined templates. The most recent work~\cite{vinod2024food} leveraged the 3D food mesh as the template to align both camera pose and object pose for portion size estimation. 

\noindent\textbf{Depth Camera-Based Approach.} The depth camera is utilized to produce a depth map that captures the distance from the camera to the food in the image. In \cite{8868629, 10220023}, the depth map is used to form a voxel representation of the image, which is then used to estimate the food volume. The main limitation is the requirement for high-quality depth maps and additional post-processing needed for consumer depth sensors.

\noindent\textbf{Deep Learning Approach.} Neural network-based methods leverage the abundance of image data to train complex networks for food portion estimation. Regression networks are used in \cite{9733557, 9874714} to estimate the energy value of food from a single image input and from an ``Energy Distribution Map," which maps the input image to the energy distribution of the foods in the image. In \cite{thames2021nutrition5k}, regression networks trained on input images and depth maps produce energy, mass, and macronutrient information for the food(s) in the image. Deep learning-based methods require large amounts of data for training and are generally not explainable. Their performance often degrades when the input test image differs significantly from the training data. 

Although these approaches have made significant strides in food portion estimation, they all face limitations that hinder their widespread adoption and accuracy in real world scenarios. Stereo-based methods are impractical for single-image input, model-based approaches struggle with diverse food shapes, depth camera-based methods require specialized hardware, and deep learning approaches lack explainability and struggle with out-of-distribution samples. To address these challenges, 3D reconstruction offers a promising solution by providing comprehensive spatial information, adapting to various food shapes, potentially working with single images, offering visually interpretable results, and enabling a standardized approach to food portion estimation. These advantages motivated the organization of the 3D Food Reconstruction challenge, with the aim of overcoming existing limitations and developing more accurate, user-friendly, and widely applicable food portion estimation techniques that can significantly impact nutritional assessment and dietary monitoring.


%% file: sec/2_Dataset.tex
\section{Datasets and Evaluation Pipeline}
\label{sec: dataset}
\subsection{Dataset Description}

The MetaFood Challenge dataset comprises 20 carefully selected food items from MetaFood3D dataset~\footnote{\href{https://lorenz.ecn.purdue.edu/~food3d/}{MetaFood3D - the dataset can be accessed at this link.}}, each scanned with a 3D scanner and accompanied by corresponding video captures. To ensure accurate size representation in the reconstructed 3D models, each item was captured alongside a checkerboard and pattern mat, serving as physical references for scaling. The challenge is structured into three difficulty levels, determined by the number of 2D images available for reconstruction:

\begin{itemize}
    \item Easy: Approximately 200 images sampled from video
    \item Medium: 30 images
    \item Hard: A single monocular top-view image
\end{itemize}

Table \ref{tab:dataset} provides detailed information about the food items in the dataset.

\begin{table}[h]
\centering
\scalebox{0.75}{
\begin{tabular}{cccc}
\hline
Object Index & Food Item & Difficulty Level & Number of Frames \\
\hline
1 & Strawberry & Easy & 199 \\
2 & Cinnamon bun & Easy & 200 \\
3 & Pork rib & Easy & 200 \\
4 & Corn & Easy & 200 \\
5 & French toast & Easy & 200 \\
6 & Sandwich & Easy & 200 \\
7 & Burger & Easy & 200 \\
8 & Cake & Easy & 200 \\
9 & Blueberry muffin & Medium & 30 \\
10 & Banana & Medium & 30 \\
11 & Salmon & Medium & 30 \\
12 & Steak & Medium & 30 \\
13 & Burrito & Medium & 30 \\
14 & Hotdog & Medium & 30 \\
15 & Chicken nugget & Medium & 30 \\
16 & Everything bagel & Hard & 1 \\
17 & Croissant & Hard & 1 \\
18 & Shrimp & Hard & 1 \\
19 & Waffle & Hard & 1 \\
20 & Pizza & Hard & 1 \\
\hline
\end{tabular}
}
\caption{MetaFood Challenge Data Details}
\label{tab:dataset}
\end{table}

\subsection{Evaluation Pipeline}

The evaluation process consists of two phases, focusing on the precision of the reconstructed 3D models in terms of their shape (3D structure) and portion size (volume).

\subsubsection{Phase-I: Volume Accuracy}

In the first phase, we employ Mean Absolute Percentage Error (MAPE) as the metric to assess the accuracy of portion size. The MAPE is calculated as follows:

\begin{equation}
    MAPE = \frac{1}{n} \sum_{i=1}^{n} \left| \frac{A_i - F_i}{A_i} \right| \times 100\%
\end{equation}

where $A_i$ is the groundtruth volume (in unit of ml) of the $i$-th food object obtained from the scanned 3D food mesh, and $F_i$ is the volume obtained from the reconstructed 3D mesh. 

\subsubsection{Phase-II: Shape Accuracy}

The top-ranking teams from Phase-I are invited to submit complete 3D mesh files for each food item. This phase involves several steps to ensure accuracy and fairness:
\begin{itemize}
    \item Model Verification: We verify the submitted models against the final Phase-I submissions to ensure consistency. Additionally, we conduct visual inspections to prevent rule violations, such as submitting primitive objects (e.g., spheres) instead of detailed reconstructions.
    \item Model Alignment: We will provide participants with the ground truth 3D models, along with the script used to compute the final Chamfer distance. Participants are required to align their models with the ground truth and prepare a transform matrix for each submitted object. The final Chamfer distance score will be computed using these submitted models and transformation matrices.
    \item Chamfer Distance Calculation: The shape accuracy is evaluated using the Chamfer distance metric. Given two point sets $X$ and $Y$, the Chamfer distance is defined as in the following. 
\end{itemize}
\begin{equation}
    d_{CD}(X,Y) = \frac{1}{|X|} \sum_{x \in X} \min_{y \in Y} \|x-y\|_2^2 + \frac{1}{|Y|} \sum_{y \in Y} \min_{x \in X} \|x-y\|_2^2
\end{equation}
This metric provides a comprehensive measure of the similarity between the reconstructed 3D models and the ground truth. 
The final ranking will be determined by combining the scores from both Phase-I (volume accuracy) and Phase-II (shape accuracy). Note that after the Phase-I evaluation, we observed some quality issues with the provided data for object 12 (steak) and object 15 (chicken nugget). To ensure the quality and fairness of the competition, we have decided to exclude these two items from the final overall evaluation process.

%% file: sec/VolETA.tex
\section{First Place Team - VolETA}
\label{sec:VolETA}

\subsection{Methodology}

The team's study utilizes multi-view reconstruction to create intricate food meshes and calculate precise food volumes. The source code is available at \url{https://github.com/GCVCG/VolETA-MetaFood}. 
\begin{figure*}[h]
    \centering
    \includegraphics[trim={0cm 0cm 0cm 2cm},clip,width=1\linewidth]{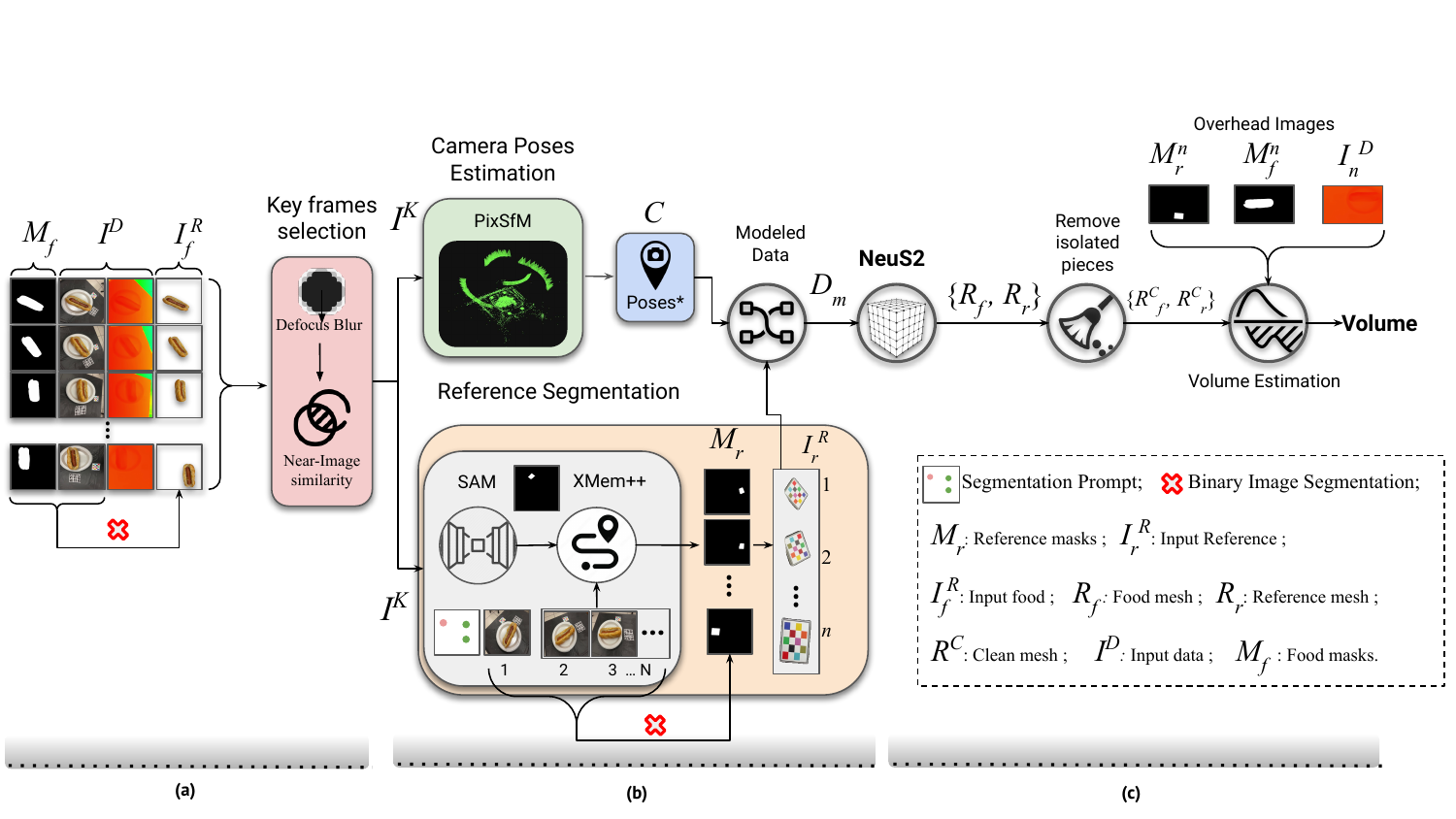}
    \caption{
The team's few-shot approach for estimating food volume in (a) a few shots involves taking (\(\mathcal{I}^D\)) and food object masks as input. The team starts by selecting keyframes based on the RGB images, removing blurry and highly overlapped images resulting (\(I^K\)). Then, (b) the team uses PixSfM to estimate camera poses (\(C\)). Simultaneously, the team segments the reference object using SAM with a segmentation prompt provided by a user. The team then uses the XMem++ method for memory-tracking to produce reference object masks for all frames, using the reference object mask and RGB images. After that, the team applies a binary image segmentation method to RGB images (\(I^K\)), reference object masks (\(M_r\)), and food object masks (\(M_f\)), resulting in RGBA images (\(I^R_r\)). In contrast, the team transforms the RGBA images and poses to generate meaningful metadata and create modeled data (\(D_m\)). Next, (c) the team inputs the modeled data into NeuS2 to reconstruct colorful meshes for reference (\(R_r\)) and food objects (\(R_f\)). To ensure accuracy, the team uses ``Remove Isolated Pieces" with diameter thresholding to clean up the mesh and remove small isolated pieces that do not belong to the reference or food mesh resulting (\(\{R^C_r, R^C_f\}\)). Finally, the team manually identifies the scaling factor using the reference mesh via MeshLab (\(S\)). The team fine-tunes the scaling factor using depth information and the food masks and then applies the fine-tuned scaling factor (\(S_f\)) to the cleaned food mesh to generate a scaled food mesh (\(R^F_f\)) in meter unit.
}
    \label{fig:methodology}
\end{figure*}

\subsubsection{Overview}
The team's approach combines computer vision and deep learning techniques to estimate food volume from RGBD images and masks accurately. Keyframe selection ensures data quality, aided by perceptual hashing and blur detection. Camera pose estimation and object segmentation lay the groundwork for neural surface reconstruction, producing detailed meshes for volume estimation. Refinement steps enhance accuracy, including isolated piece removal and scaling factor adjustment. The team's approach offers a comprehensive solution for precise food volume assessment with potential applications in nutrition analysis.

\subsubsection{The Team's Proposal: VolETA}
The team begins their approach by acquiring input data, specifically RGBD images and corresponding food object masks. These RGBD images, denoted as $\mathcal{I}^D=\{{I_i^D}\}_{i=1}^n$, where \( n \) is the total number of frames, provide the necessary depth information alongside the RGB images. The food object masks, denoted as \( \{M_f^i\}_{i=1}^n \), aid in identifying the regions of interest within these images.

Next, the team proceeds with keyframe selection. From the set \( \{I^D_i\}_{i=1}^n \), keyframes \( \{I_i^K\}_{j=1}^k \subseteq \{I^D_i\}_{i=1}^n \) are selected. The team implements a method to detect and remove duplicates \cite{idealods2019imagededup} and blurry images \cite{de2013image} to ensure high-quality frames. This involves applying the Gaussian blurring kernel followed by the fast Fourier transform method. Near-Image Similarity \cite{idealods2019imagededup} employs a perceptual hashing and hamming distance thresholding to detect similar images and keep overlapping. The duplicates and blurry images are excluded from the selection process to maintain data integrity and accuracy, as shown in Fig.~\ref{fig:methodology}(a).

Using the selected keyframes \( \{I^K_j\}_{j=1}^k \), the team estimates the camera poses through PixSfM \cite{lindenberger2021pixel} (i.e., extracting features using SuperPoint \cite{detone2018superpoint}, matching them using SuperGlue \cite{sarlin2020superglue}, and refining them). The outputs are the set of camera poses \( \{C_j\}_{j=1}^k \), which are crucial for spatial understanding of the scene.

In parallel, the team utilizes the SAM \cite{kirillov2023segment} for reference object segmentation. SAM segments the reference object with a user-provided segmentation prompt (i.e., user click), producing a reference object mask \( M^R \) for each keyframe. This mask is a foundation for tracking the reference object across all frames. The team then applies the XMem++ \cite{bekuzarov2023xmem++} method for memory tracking, which extends the reference object mask \( M^R \) to all frames, resulting in a comprehensive set of reference object masks \( \{M^R_i\}_{i=1}^n \). This ensures consistency in reference object identification throughout the dataset.

To create RGBA images, the team combines the RGB images, reference object masks \( \{M^R_i\}_{i=1}^n \), and food object masks \( \{M^F_i\}_{i=1}^n \). This step, denoted as \( \{I^R_i\}_{i=1}^n \), integrates the various data sources into a unified format suitable for further processing, as shown in Fig.~\ref{fig:methodology}(b).

The team converts the RGBA images \( \{I^R_i\}_{i=1}^n \) and camera poses \( \{C_j\}_{j=1}^k \) into meaningful metadata and modeled data \( D_m \). This transformation facilitates the accurate reconstruction of the scene. 

The modeled data \( D_m \) is then input into NeuS2 \cite{wang2023neus2} for mesh reconstruction. NeuS2 generates colorful meshes \( \{R_f, R_r\} \) for the reference and food objects, providing detailed 3D representations of the scene components. The team applies the ``Remove Isolated Pieces" technique to refine the reconstructed meshes. Given that the scenes contain only one food item, the team sets the diameter threshold to 5\% of the mesh size. This method deletes isolated connected components whose diameter is less than or equal to this 5\% threshold, resulting in a cleaned mesh \( \{R^C_f, R^C_r\} \). This step ensures that only significant and relevant parts of the mesh are retained.

The team manually identifies an initial scaling factor \( S \) using the reference mesh via MeshLab \cite{cignoni2008meshlab} for scaling factor identification. This factor is then fine-tuned \( S_{f} \) using depth information and food and reference masks, ensuring accurate scaling relative to real-world dimensions. Finally, the fine-tuned scaling factor \( S_{f} \) is applied to the cleaned food mesh \( R^C_f \), producing the final scaled food mesh \( R^F_f \). This step culminates in an accurately scaled 3D representation of the food object, enabling precise volume estimation, as shown in Fig.~\ref{fig:methodology}(c).

\subsubsection{Detecting the scaling factor}
\begin{figure}[htb]
    \centering
    \begin{subfigure}[b]{0.45\linewidth}
         \centering
            \includegraphics[width=1\linewidth, angle=90]{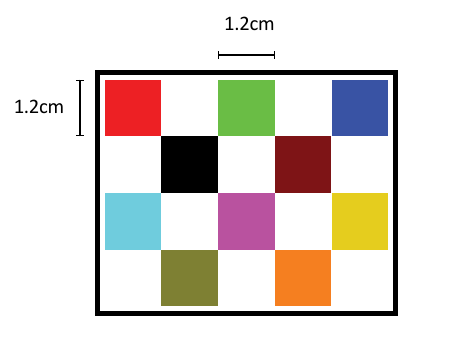}
         \caption{ }
         \label{fig:chessboard}
     \end{subfigure}
    \begin{subfigure}[b]{0.45\linewidth}
         \centering
         \includegraphics[width=1\linewidth]{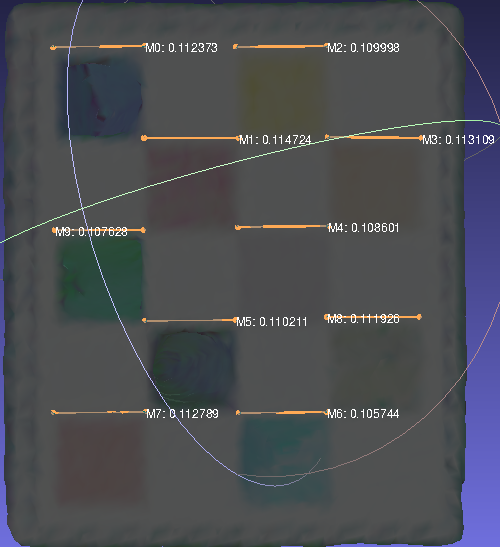}
         \caption{ }
         \label{fig:meshlab_scaling_factor}
     \end{subfigure}
    \caption{The team manually measures the scaling factor using MeshLab's Measuring tool. The team measures multiple blocks in the reference object mesh; then, the team takes the average of blocks lengths \(l_{avg}\). }
    \label{fig:scaling_factor}
\end{figure}

Generally, 3D reconstruction methods generate unitless meshes (i.e., no physical scale) by default. To overcome this limitation, the team manually identifies the scaling factor by measuring the distance for each block for the reference object mesh, as shown in Fig.~\ref{fig:meshlab_scaling_factor}. Next, the team takes the average of all blocks lengths \(l_{avg}\), while the actual real-world length (as shown in Fig.~\ref{fig:chessboard}) is constant \(l_{real}=0.012\) in meter. Furthermore, the team applies the scaling factor \(S=l_{real}/l_{avg}\) on the clean food mesh \( R^C_f \), producing the final scaled food mesh \( R^F_f \) in meter.

The team leverages depth information alongside food and reference object masks to validate the scaling factors. The team's method for assessing food size entails utilizing overhead RGB images for each scene. Initially, the team determines the pixel-per-unit (PPU) ratio (in meters) using the reference object. Subsequently, the team extracts the food width (\(f_{w}\)) and length (\(f_{l}\)) employing a food object mask. To ascertain the food height (\(f_{h}\)), the team follows a two-step process. Firstly, the team conducts binary image segmentation using the overhead depth and reference images, yielding a segmented depth image for the reference object. The team then calculates the average depth utilizing the segmented reference object depth (\(d_{r}\)). Similarly, employing binary image segmentation with an overhead food object mask and depth image, the team computes the average depth for the segmented food depth image (\(d_{f}\)). Finally, the estimated food height \(f_{h}\) is computed as the absolute difference between \(d_{r}\) and \(d_{f}\). Furthermore, to assess the accuracy of the scaling factor \(S\), the team computes the food bounding box volume (\((f_{w} \times f_{l} \times f_{h}) \times PPU \)). The team evaluates if the scaling factor \(S\) generates a food volume close to this potential volume, resulting in \(S_{fine}\). Table~\ref{tab:scaling_factors} shows the scaling factors, PPU, 2D Reference object dimensions, 3D food object dimensions, and potential volume.

\begin{table*}[h]
    \small
    \centering
    \begin{tabular}{c|c|l|c|c|cc|ccc|c}
    \hline
         Level & Id & Label & \(S_{f}\) & PPU & \multicolumn{2}{c|}{\(R_{w} \times R_{l}\)} &  \multicolumn{3}{c|}{(\(f_{w} \times f_{l} \times f_{h}\))} & Volume (\(cm^3\))   
         \\ \hline
\multirow{8}{*}{Easy} & 1 & strawberry\_2 & 0.08955223881 & 0.01786 & 320 & 360 & 238 & 257 & 2.353 & 45.91\\
& 2 & cinnamon\_bun\_1 & 0.1043478261 & 0.02347 & 236 & 274 & 363 & 419 & 2.353 & 197.07\\
& 3 & pork\_rib\_2 & 0.1043478261 & 0.02381 & 246 & 270 & 435 & 778 & 1.176 & 225.79\\
& 4 & corn\_2 & 0.08823529412 & 0.01897 & 291 & 339 & 262 & 976 & 2.353 & 216.45\\
& 5 & french\_toast\_2 & 0.1034482759 & 0.02202 & 266 & 292 & 530 & 581 & 2.53 & 377.66\\
& 6 & sandwich\_2 & 0.1276595745 & 0.02426 & 230 & 265 & 294 & 431 & 2.353 & 175.52\\
& 7 & burger\_1 & 0.1043478261 & 0.02435 & 208 & 264 & 378 & 400 & 2.353 & 211.03\\
& 8 & cake\_1 & 0.1276595745 & 0.02143 & 256 & 300 & 298 & 310 & 4.706 & 199.69\\
\hline
\multirow{7}{*}{Medium} & 9 & blueberry\_muffin & 0.08759124088 & 0.01801 & 291 & 357 & 441 & 443 & 2.353 & 149.12 \\
& 10 & banana\_2 & 0.08759124088 & 0.01705 & 315 & 377 & 446 & 857 & 1.176 & 130.80\\
& 11 & salmon\_1 & 0.1043478261 & 0.02390 & 242 & 269 & 201 & 303 & 1.176 & 40.94\\
& 13 & burrito\_1 & 0.1034482759 & 0.02372 & 244 & 271 & 251 & 917 & 2.353 & 304.87\\
& 14 & frankfurt\_sandwich\_2 & 0.1034482759 & 0.02115 & 266 & 304 & 400 & 1022 & 2.353 & 430.29\\
\hline
\multirow{5}{*}{Hard} & 16 & everything\_bagel & 0.08759124088 & 0.01747 & 306 & 368 & 458 & 484 & 1.176 & 79.61\\
& 17 & croissant\_2 & 0.1276595745 & 0.01751 & 319 & 367 & 395 & 695 & 2.176 & 183.39 \\
& 18 & shrimp\_2 & 0.08759124088 & 0.02021 & 249 & 318 & 186 & 195 & 0.987 & 14.64\\
& 19 & waffle\_2 & 0.01034482759 & 0.01902 & 294 & 338 & 465 & 537 & 0.8 & 72.29 \\ & 20 & pizza & 0.01034482759 & 0.01913 & 292 & 336 & 442 & 651 & 1.176 & 123.97 \\
    \hline
    \end{tabular}
    \caption{A list of information that extracted using the RGBD and masks, where the team presents the scene Id, the scaling factor \(S_{fine}\), Pixel-Per-Unit (in cm), 2D reference object dimensions ($R_w \times R_l$), 3D food object dimensions ($f_w\times f_l \times f_h$) in pixels, and the potential volume (in $cm^3$). The rows in red are excluded meshes.}
    \label{tab:scaling_factors}
\end{table*}

\begin{figure}[htb]
    \centering
    \includegraphics[trim={6cm 4cm 5cm 3.5cm},clip,width=1\linewidth]{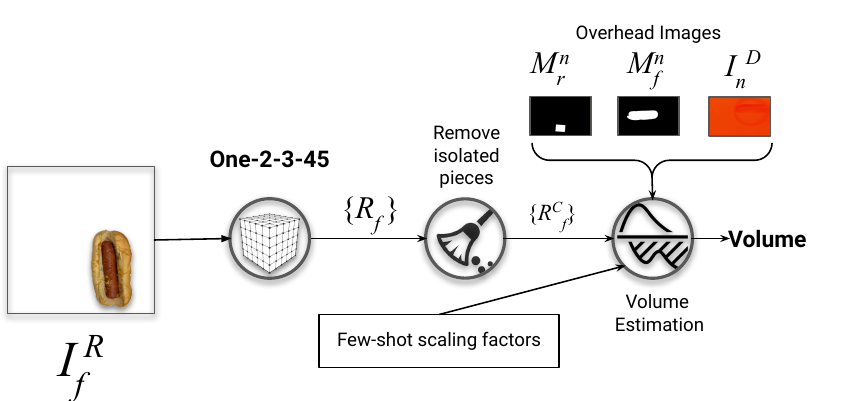}
    \caption{The team's one-shot food volume estimation approach. The team begins with a food-segmented image (\(I^R_f\)), and then uses the One-2-3-45 model to generate a mesh (\(R_f\)). Next, the team cleans up the isolated pieces that are less than 5\% of the (\(R_f\)) size, resulting in a cleaned food mesh \(R^C_f\). Furthermore, the team chooses a scaling factor based on the depth information \(S_f\). Finally, the team applies the chosen scaling factor on \(R^C_f\) to have a scaled mesh (\(R^F_f\)) where the team extracts the volume.}
    \label{fig:one-shot-methodology}
\end{figure}

For one-shot 3D reconstruction, the team leverages One-2-3-45 \cite{liu2024one} for reconstructing a 3D from a single RGBA view input after applying binary image segmentation on both food RGB and mask. Next, the team removes isolated pieces from the generated mesh. After that, the team reuses the scaling factor \(S\), which is closer to the potential volume of the clean mesh, as shown in Fig.~\ref{fig:one-shot-methodology}.

\subsection{Experimental Results}
\label{sec:results}

\begin{figure*}[h]
    \centering
    \begin{subfigure}[b]{0.16\linewidth}
         \centering
            \includegraphics[trim={5cm 2cm 5cm 3cm},clip,width=0.45\linewidth]{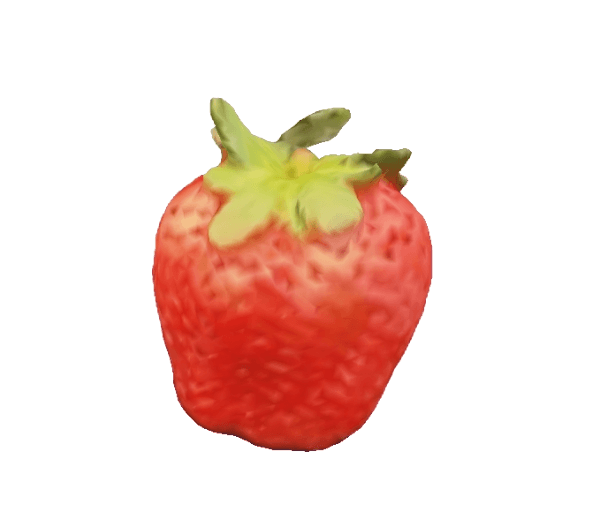}
            \includegraphics[trim={5cm 2cm 5cm 3cm},clip,width=0.45\linewidth]{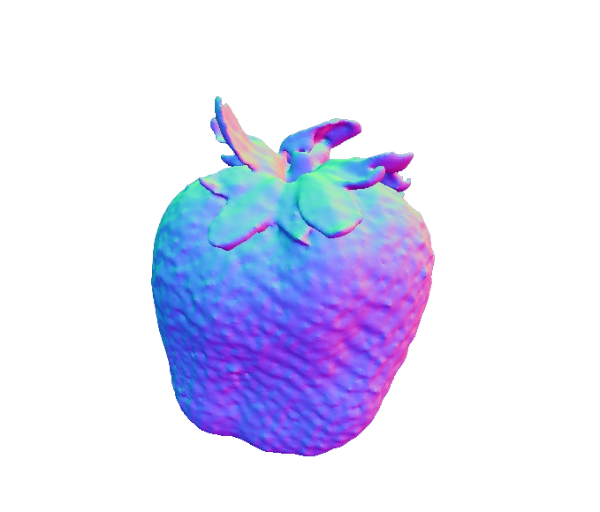}
         \caption{strawberry (1)}
         \label{fig:rsl_1}
     \end{subfigure}
     \begin{subfigure}[b]{0.16\linewidth}
         \centering
            \includegraphics[trim={5cm 5cm 5cm 5cm},clip,width=0.45\linewidth]{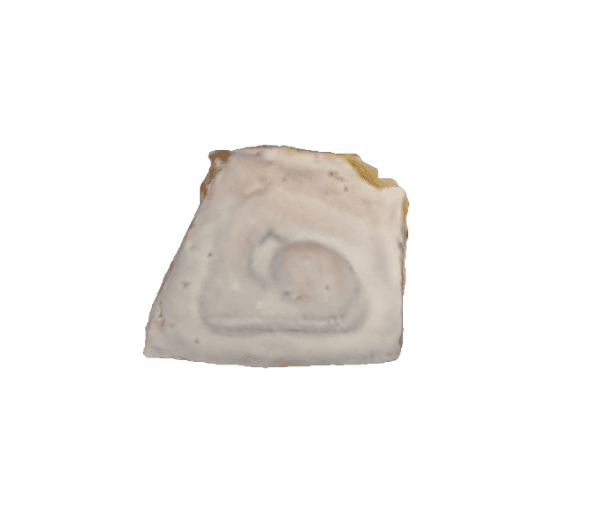}
            \includegraphics[trim={4cm 5cm 5cm 5cm},clip,width=0.45\linewidth]{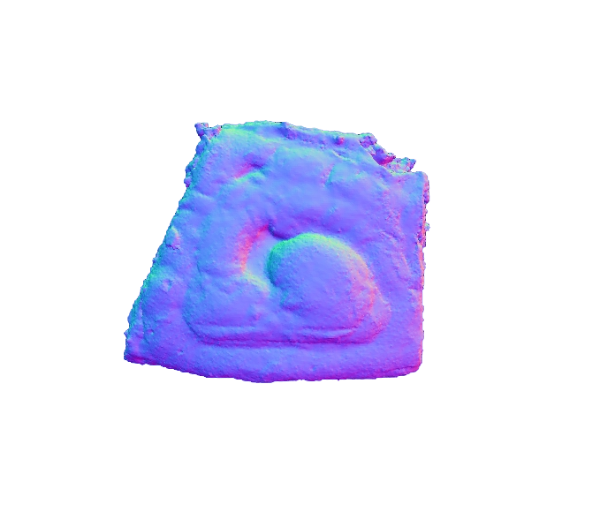}
         \caption{cinnamon bun (2)}
         \label{fig:rsl_2}
     \end{subfigure}
     \begin{subfigure}[b]{0.16\linewidth}
         \centering
            \includegraphics[trim={5cm 3cm 5cm 3cm},clip,width=0.45\linewidth]{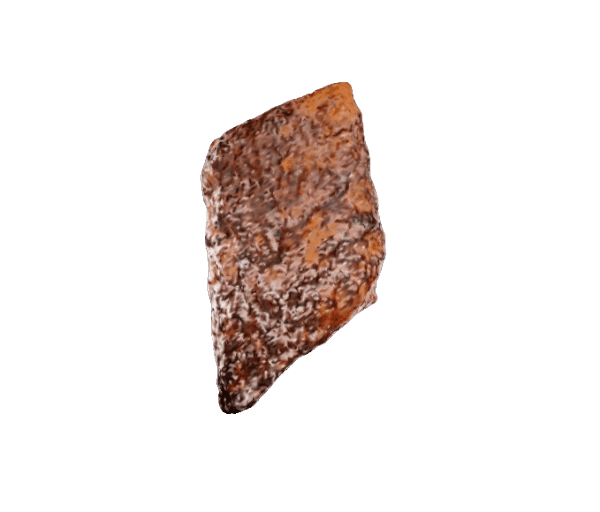}
            \includegraphics[trim={5cm 1cm 5cm 3cm},clip,width=0.45\linewidth]{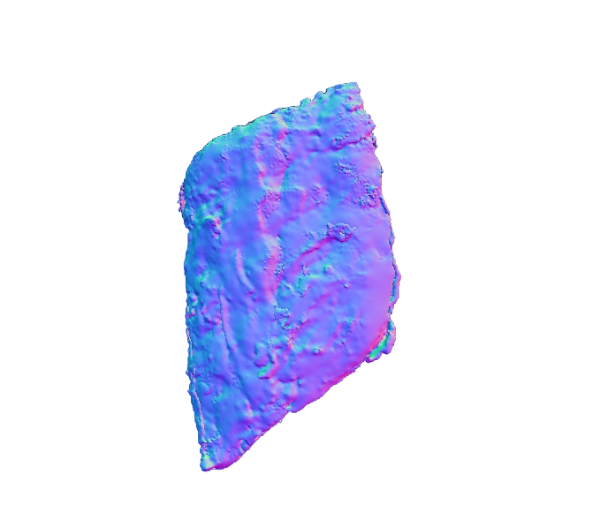}
         \caption{pork rib (3)}
         \label{fig:rsl_3}
     \end{subfigure}
      \begin{subfigure}[b]{0.16\linewidth}
         \centering
            \includegraphics[trim={5cm 3cm 5cm 3cm},clip,width=0.45\linewidth]{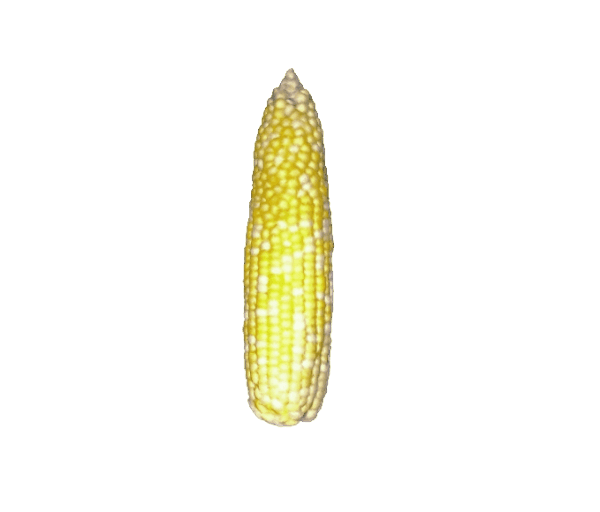}
            \includegraphics[trim={5cm 3cm 5cm 3cm},clip,width=0.45\linewidth]{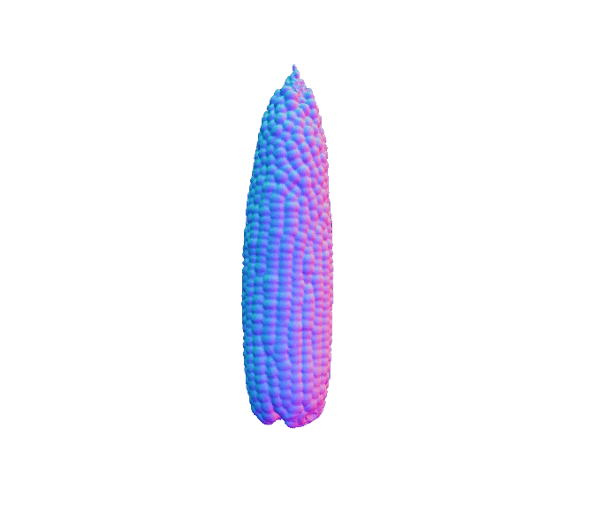}
         \caption{corn (4)}
         \label{fig:rsl_4}
     \end{subfigure}
     \begin{subfigure}[b]{0.16\linewidth}
         \centering
            \includegraphics[trim={5cm 5cm 5cm 5cm},clip,width=0.45\linewidth]{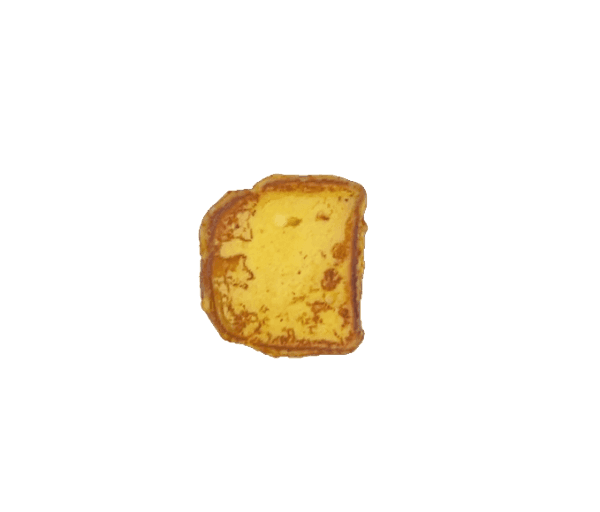}
            \includegraphics[trim={5cm 4.5cm 5cm 5cm},clip,width=0.45\linewidth]{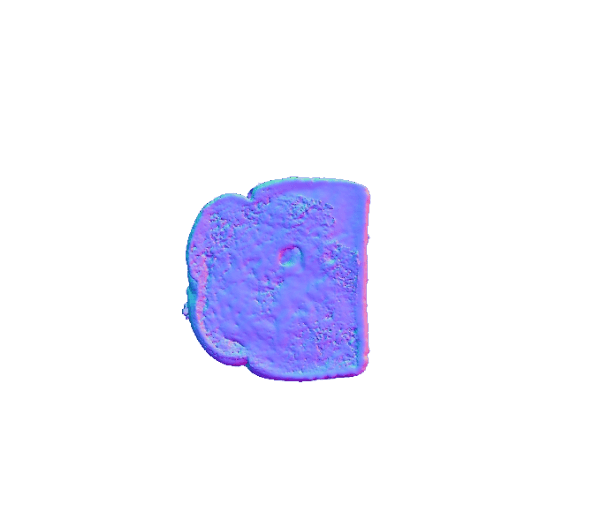}
         \caption{french toast (5)}
         \label{fig:rsl_5}
     \end{subfigure}
     \begin{subfigure}[b]{0.16\linewidth}
         \centering
            \includegraphics[trim={5cm 4cm 5cm 5cm},clip,width=0.45\linewidth]{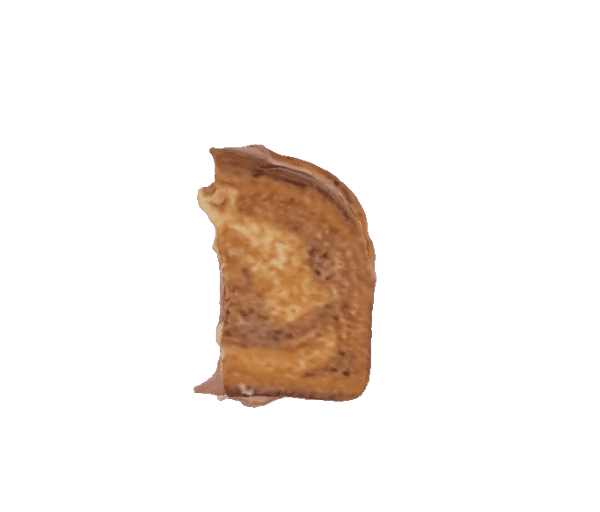}
            \includegraphics[trim={5cm 4cm 5cm 5cm},clip,width=0.45\linewidth]{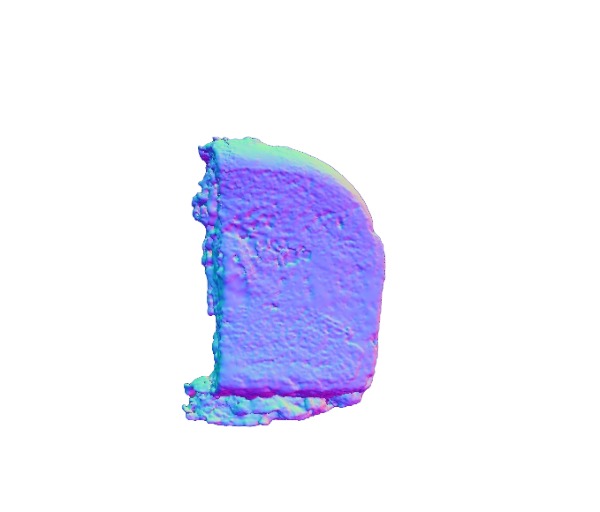}
         \caption{sandwich (6)}
         \label{fig:rsl_6}
     \end{subfigure}
     \begin{subfigure}[b]{0.16\linewidth}
         \centering
            \includegraphics[trim={6cm 5cm 6cm 6cm},clip,width=0.45\linewidth]{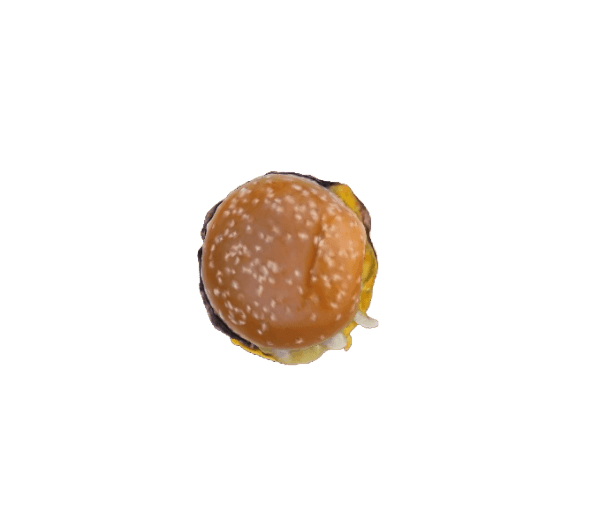}
            \includegraphics[trim={6cm 5cm 6cm 6cm},clip,width=0.45\linewidth]{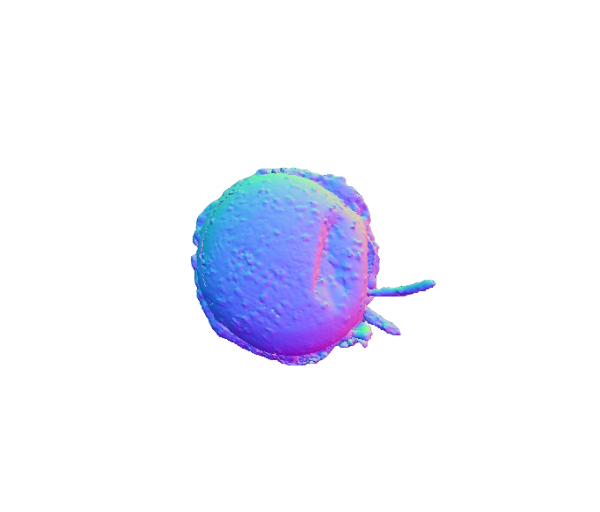}
         \caption{burger (7)}
         \label{fig:rsl_7}
     \end{subfigure}
      \begin{subfigure}[b]{0.16\linewidth}
         \centering
            \includegraphics[trim={6cm 6cm 6cm 6cm},clip,width=0.45\linewidth]{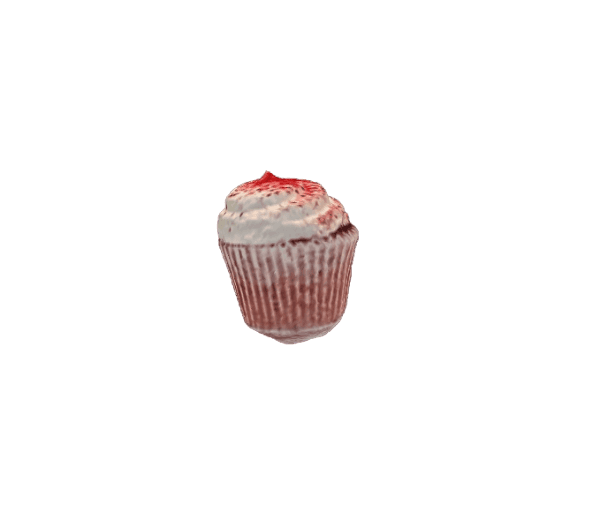}
            \includegraphics[trim={6cm 6cm 6cm 6cm},clip,width=0.45\linewidth]{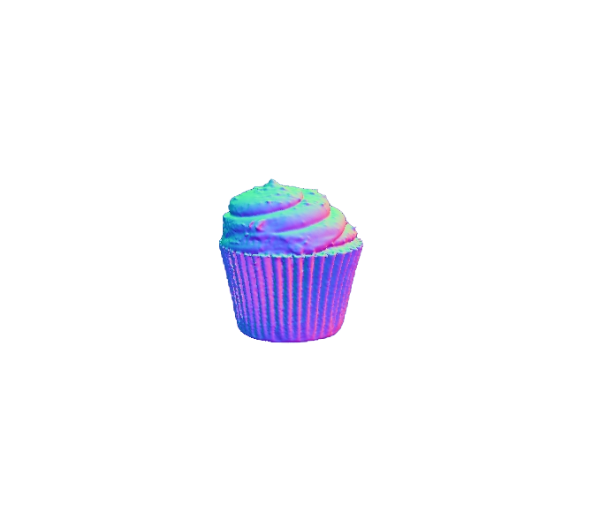}
         \caption{cake (8)}
         \label{fig:rsl_8}
     \end{subfigure}
     \begin{subfigure}[b]{0.16\linewidth}
         \centering
            \includegraphics[trim={5cm 5cm 5cm 5cm},clip,width=0.45\linewidth]{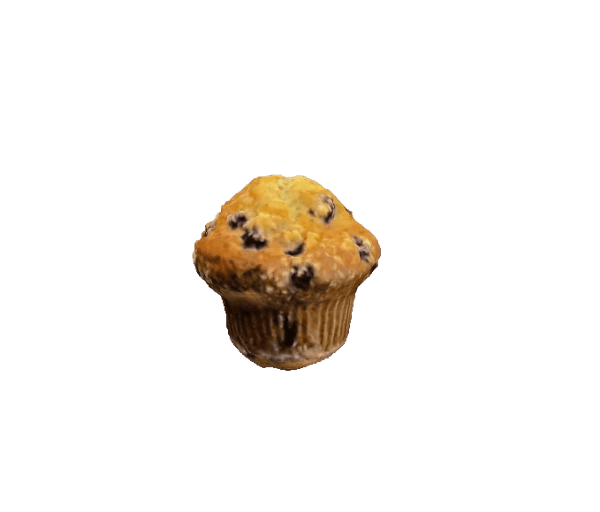}
            \includegraphics[trim={5cm 5cm 5cm 5cm},clip,width=0.45\linewidth]{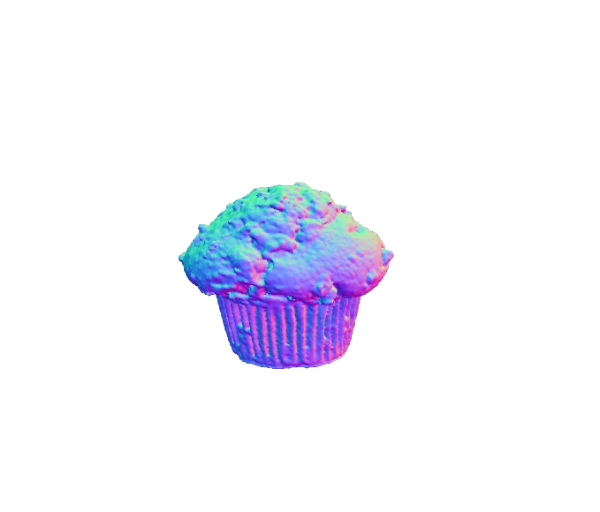}
         \caption{blueberry muffin (9)}
         \label{fig:rsl_9}
     \end{subfigure}
     \begin{subfigure}[b]{0.16\linewidth}
         \centering
            \includegraphics[trim={5cm 2cm 5cm 3cm},clip,width=0.45\linewidth]{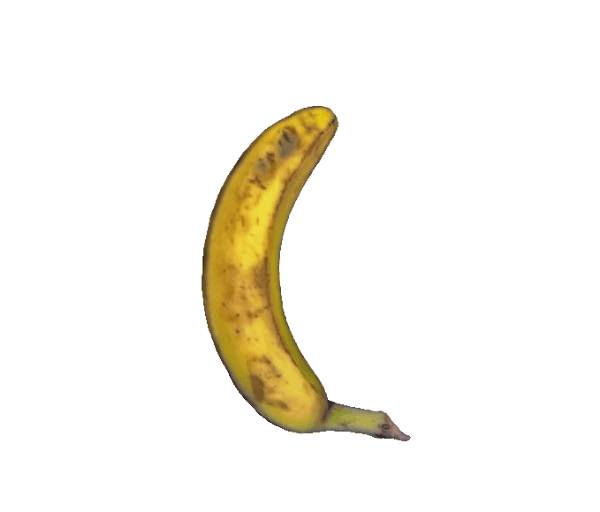}
            \includegraphics[trim={5cm 2cm 5cm 3cm},clip,width=0.45\linewidth]{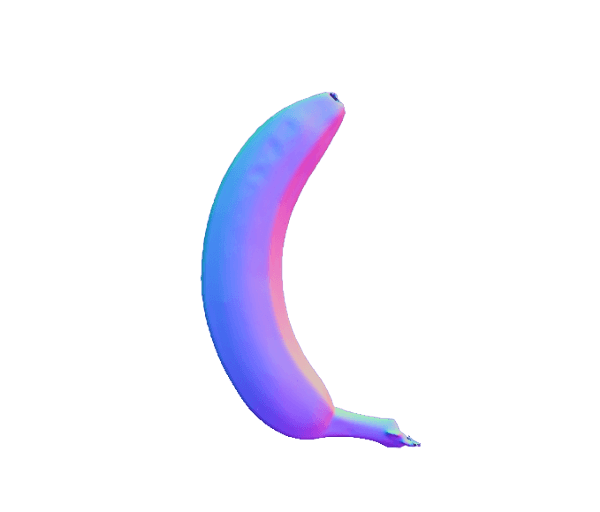}
         \caption{banana (10)}
         \label{fig:rsl_10}
     \end{subfigure}
     \begin{subfigure}[b]{0.16\linewidth}
         \centering
            \includegraphics[trim={6cm 5cm 6cm 6cm},clip,width=0.45\linewidth]{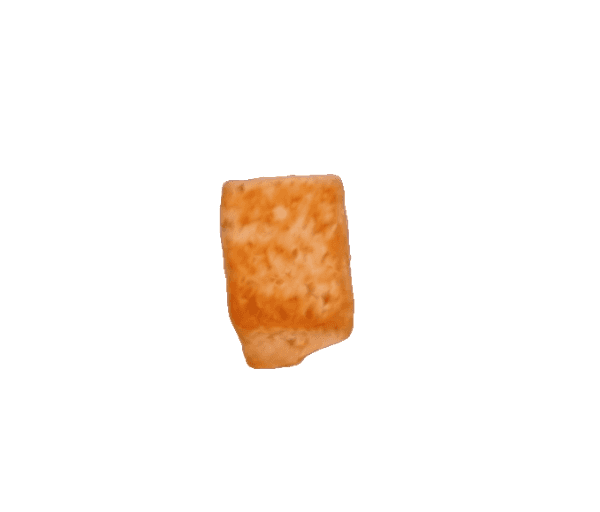}
            \includegraphics[trim={6cm 4.5cm 6cm 6cm},clip,width=0.45\linewidth]{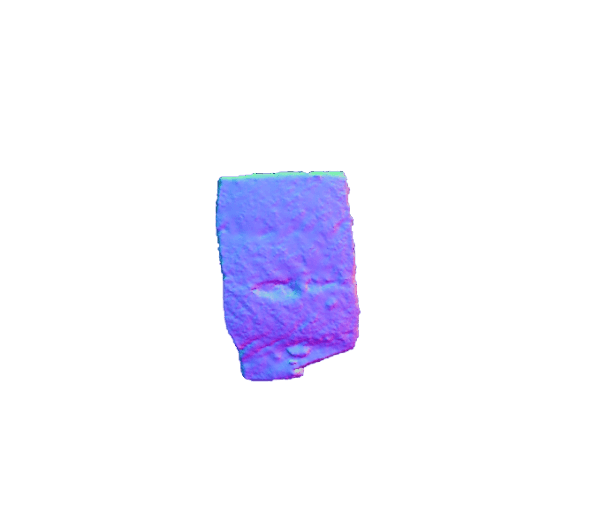}
         \caption{salmon (11)}
         \label{fig:rsl_11}
     \end{subfigure}
     \begin{subfigure}[b]{0.16\linewidth}
         \centering
            \includegraphics[trim={5cm 5cm 5cm 5cm},clip,angle=90,width=0.45\linewidth]{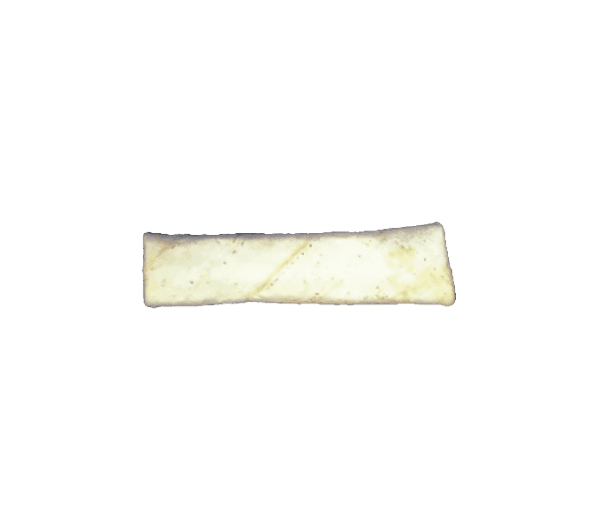}
            \includegraphics[trim={5cm 5cm 5cm 5cm},clip,angle=90,width=0.45\linewidth]{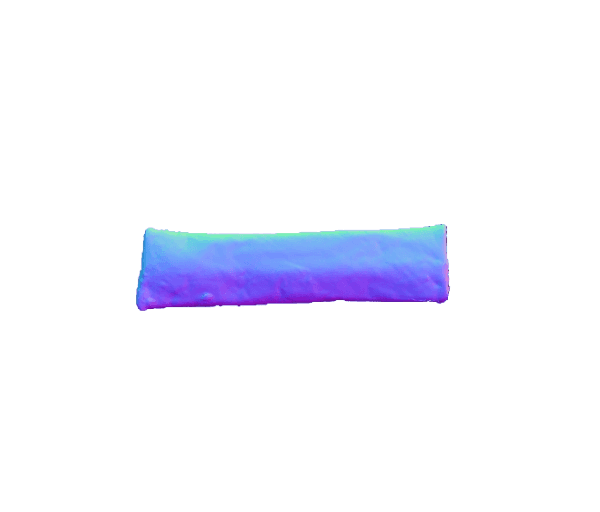}
         \caption{burrito (13)}
         \label{fig:rsl_13}
     \end{subfigure}
     \begin{subfigure}[b]{0.16\linewidth}
         \centering
            \includegraphics[trim={5cm 2cm 5cm 3cm},clip,width=0.45\linewidth]{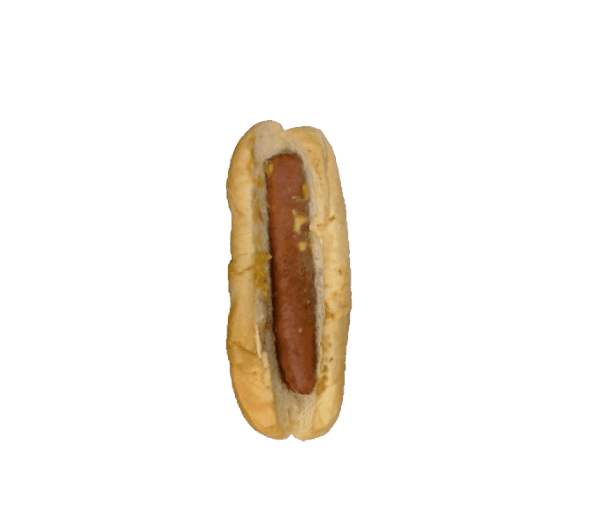}
            \includegraphics[trim={5cm 2cm 5cm 3cm},clip,width=0.45\linewidth]{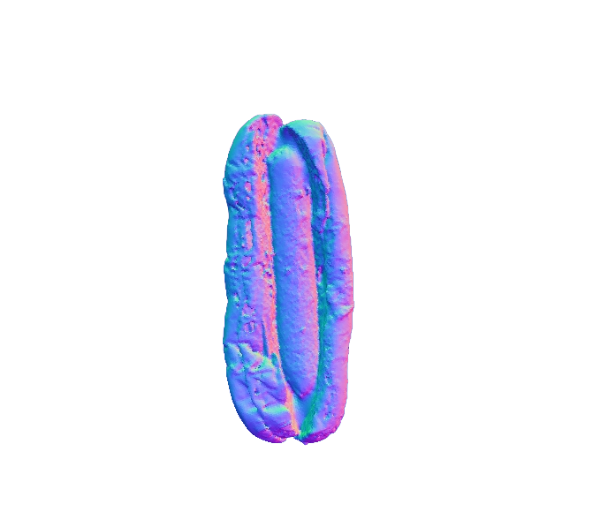}
         \caption{frank sandwich (14)}
         \label{fig:rsl_14}
     \end{subfigure}
     \begin{subfigure}[b]{0.16\linewidth}
         \centering
            \includegraphics[trim={7cm 6cm 6cm 6cm},clip,width=0.45\linewidth]{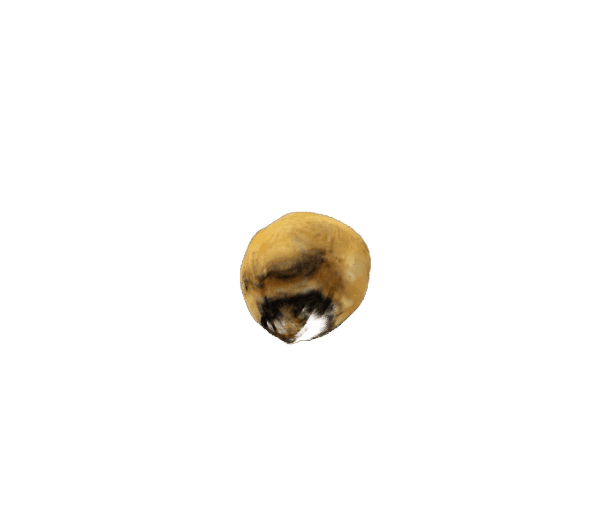}
            \includegraphics[trim={6cm 5cm 6cm 6cm},clip,width=0.45\linewidth]{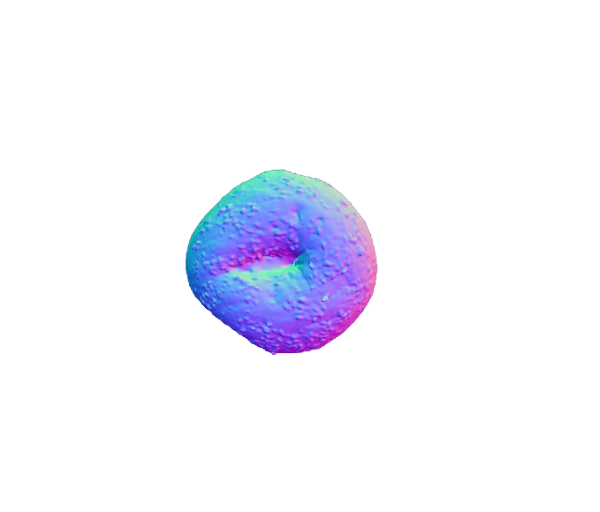}
         \caption{everything bagel (16)}
         \label{fig:rsl_16}
     \end{subfigure}
     \begin{subfigure}[b]{0.16\linewidth}
         \centering
            \includegraphics[trim={5cm 4cm 5cm 5cm},clip,width=0.45\linewidth]{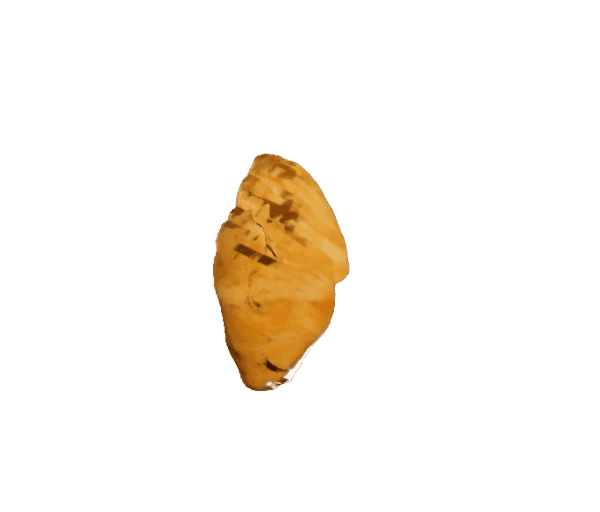}
            \includegraphics[trim={5cm 4cm 5cm 5cm},clip,width=0.45\linewidth]{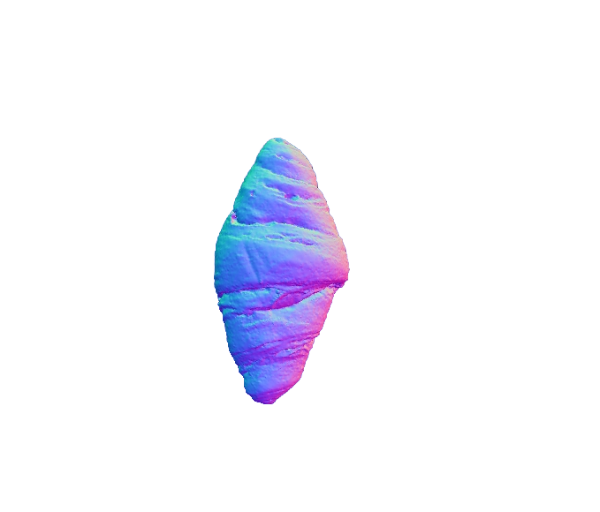}
         \caption{croissant (17)}
         \label{fig:rsl_17}
     \end{subfigure}
     \begin{subfigure}[b]{0.16\linewidth}
         \centering
            \includegraphics[trim={5cm 5cm 5cm 5cm},clip,width=0.45\linewidth]{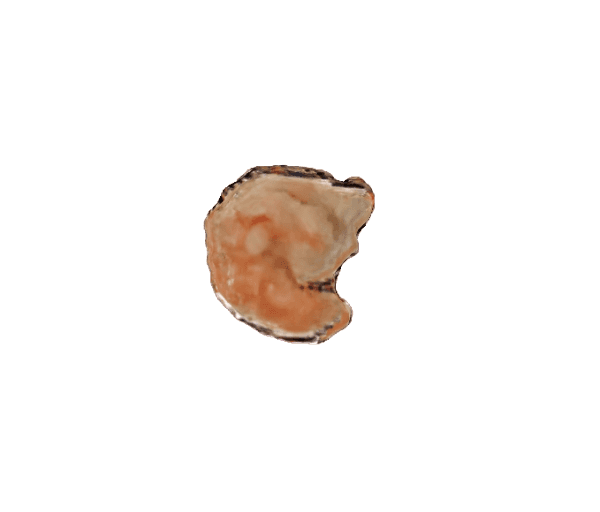}
            \includegraphics[trim={5cm 5cm 5cm 5cm},clip,width=0.45\linewidth]{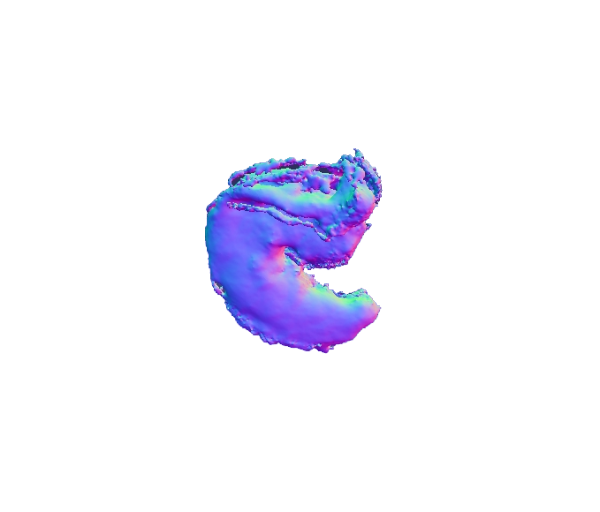}
         \caption{shrimp (18)}
         \label{fig:rsl_18}
     \end{subfigure}
     \begin{subfigure}[b]{0.16\linewidth}
         \centering
            \includegraphics[trim={5cm 5cm 5cm 5cm},clip,width=0.45\linewidth]{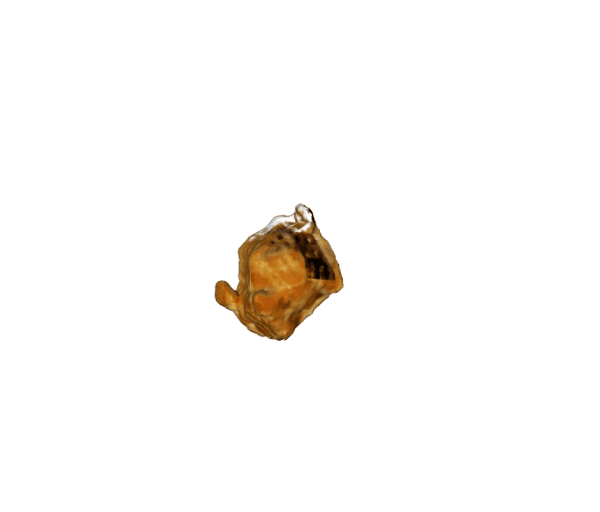}
            \includegraphics[trim={5cm 5cm 5cm 5cm},clip,width=0.45\linewidth]{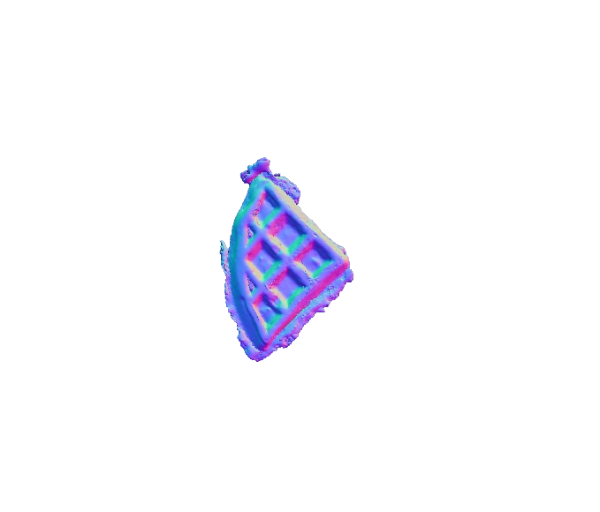}
         \caption{waffle (19)}
         \label{fig:rsl_19}
     \end{subfigure}
     \begin{subfigure}[b]{0.16\linewidth}
         \centering
            \includegraphics[trim={5cm 4cm 5cm 5cm},clip,width=0.45\linewidth]{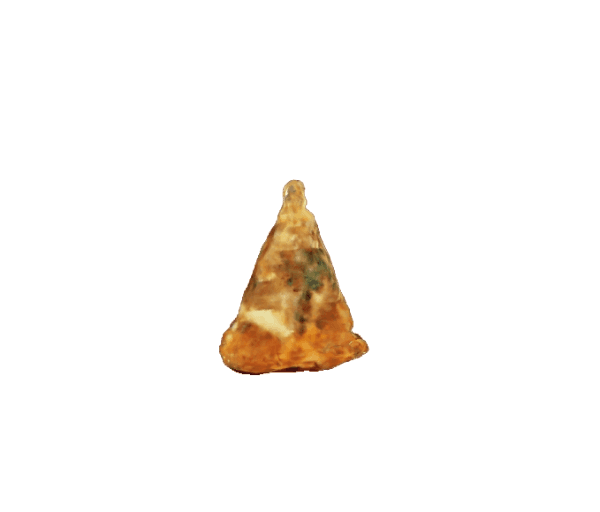}
            \includegraphics[trim={5cm 4cm 5cm 4.5cm},clip,width=0.45\linewidth]{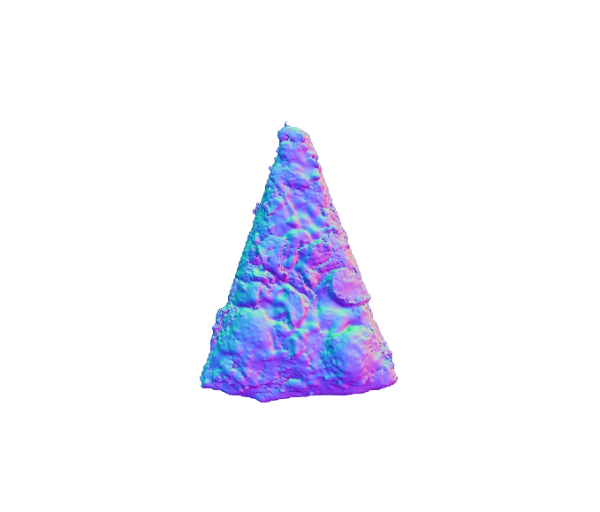}
         \caption{pizza (20)}
         \label{fig:rsl_20}
     \end{subfigure}
    \caption{
    Comparisons to the team's results and ground truth using the challenge dataset. Each scene shows the team's reconstruction (left) and ground truth (right).   
    }
    \label{fig:qualitative_results}
\end{figure*}


\subsubsection{Implementation settings}
\label{sec:resource_limitation}
The team ran the experiments using two GPUs, GeForce GTX 1080 Ti/12G and RTX 3060/6G. The team set the hamming distance as 12 for the near image similarity. For Gaussian kernel radius, the team set the even numbers in the range $[0 ... 30]$ for detecting blurry images. The team set the diameter as 5\% for removing isolated pieces. The number of iteration of NeuS2 is 15000, mesh resolution is $512 \times 512$, the unit cube ``aabb\_scale" is $1$, ``scale": 0.15, and ``offset": $[0.5, 0.5, 0.5]$ for each food scene.


\subsubsection{VolETA Results}
\begin{figure}[htb]
    \centering
    \includegraphics[width=1\linewidth]{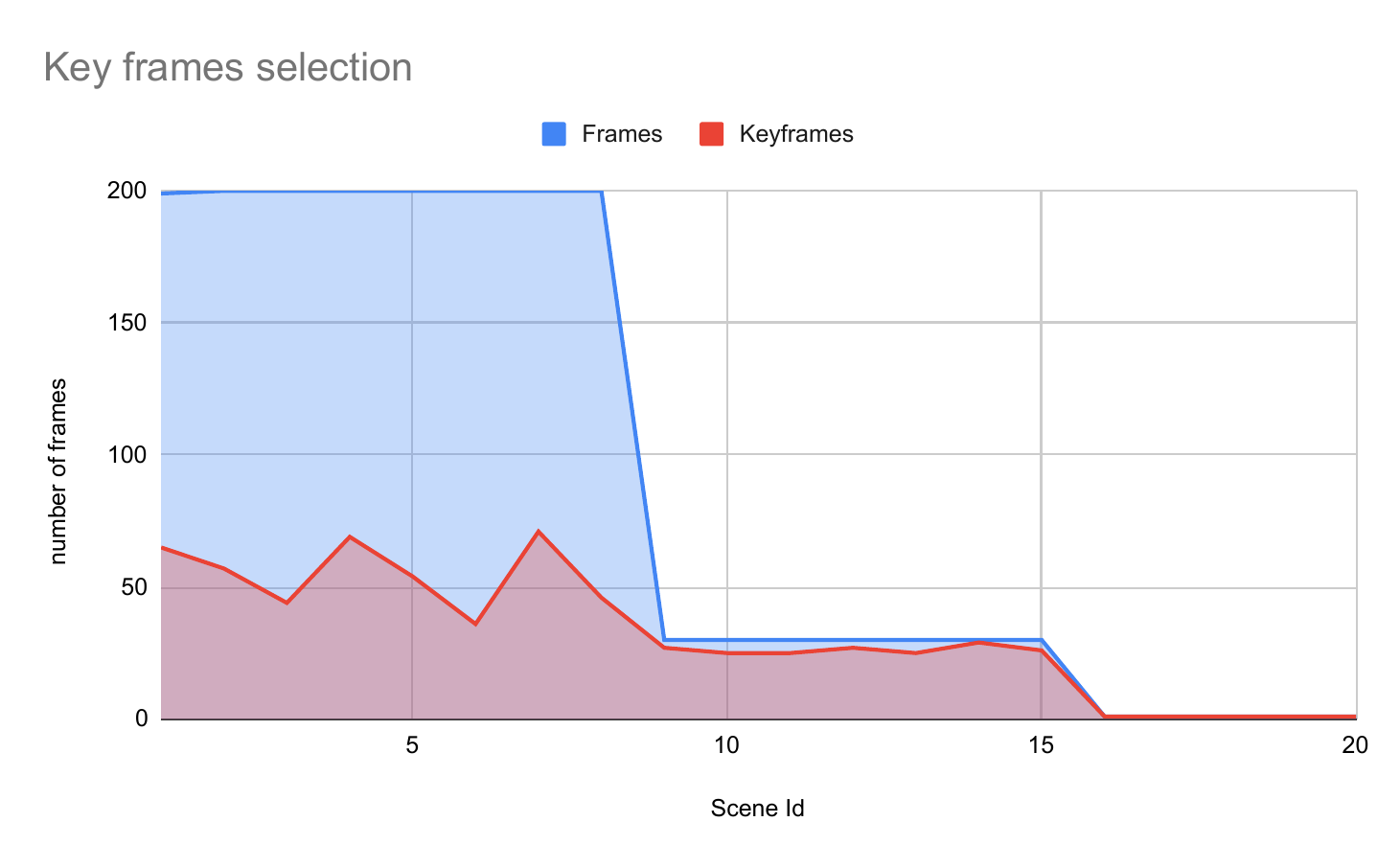}
    \caption{A quantitative results to the number of frames before and after the keyframe selection phase. The team's approach is only using 34.8\% of the data.}
    \label{fig:keyframe_selection}
\end{figure}

The team extensively validated their approach on the challenge dataset as described in Section~\ref{sec: dataset} and compared their results with ground truth meshes using MAPE and Chamfer distance metrics. More Briefly, the team leverages their approach for each food scene separately. A one-shot food volume estimation approach is applied if the number of keyframes $k$ equals 1. Otherwise, a few-shot food volume estimation is applied. Notably, Fig.~\ref{fig:keyframe_selection} shows that the team's keyframe selection process chooses 34.8\% of total frames for the rest of the pipeline, where it shows the minimum frames with the highest information.

\begin{figure}[htb]
    \centering
     \begin{subfigure}[b]{0.49\linewidth}
         \centering
            \includegraphics[width=1\linewidth]{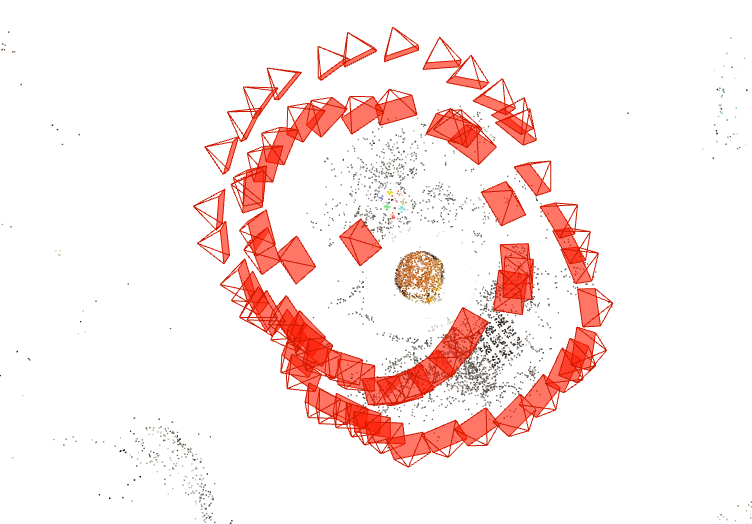}
         \caption{burger (7)}
         \label{fig:burger_1}
     \end{subfigure}
     \begin{subfigure}[b]{0.49\linewidth}
         \centering
            \includegraphics[width=1\linewidth]{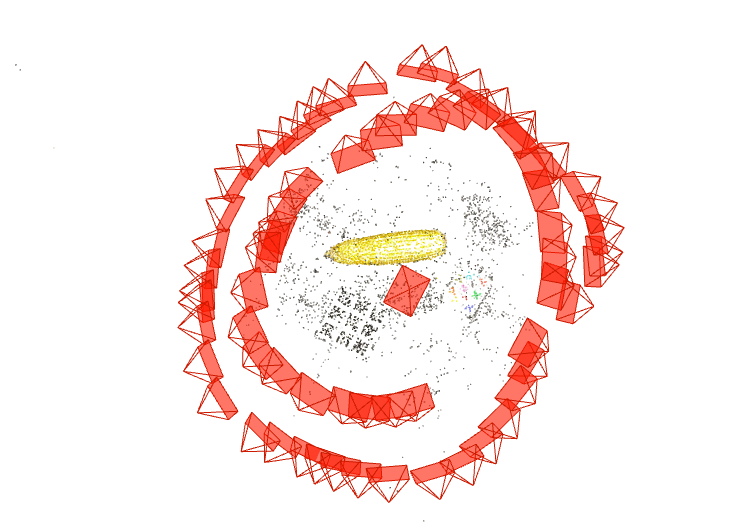}
         \caption{corn (4)}
         \label{fig:corn_2}
     \end{subfigure}
    \caption{PixSfm results after applying keyframes selection. PixSfM excels in estimating and refining camera poses by providing a rich point cloud using Superpoint feature extractors.}
    \label{fig:pixsfm}
\end{figure}

After finding the keyframes, PixSfM \cite{lindenberger2021pixel} estimates the poses and point cloud (see Fig.~\ref{fig:pixsfm}).

\begin{figure}[htb]
    \centering
        \includegraphics[width=1\linewidth]{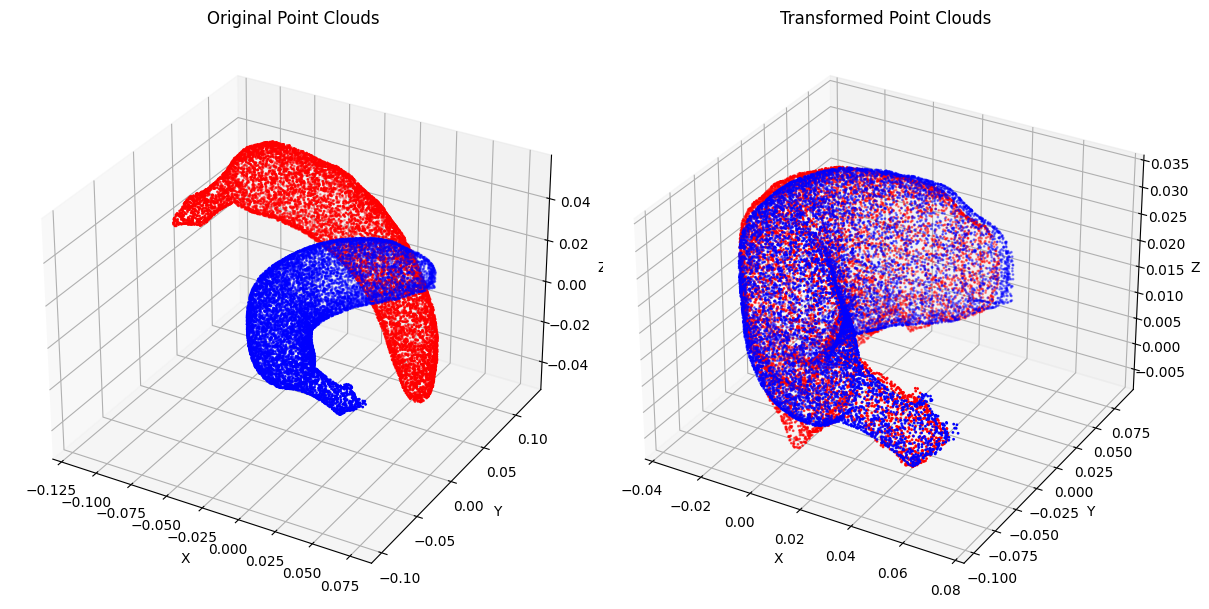}
    \caption{Performed ICP mesh registration between the team's generated and ground truth meshes for banana\_2 scene. Unregistered meshes are on the left, while registered meshes are on the right. The team's Point cloud is red, and the ground truth is blue.}
    \label{fig:icp}
\end{figure}

After generating the scaled meshes, the team calculates the volumes and Chamfer distance with and without transformation metrics. The team registered their meshes and ground truth meshes to obtain the transformation metrics using ICP \cite{rusinkiewicz2001efficient} (see Fig.~\ref{fig:icp}).

Table ~\ref{tab:results_1} presents the quantitative comparisons of the team's volumes and Chamfer distance with and without the estimated transformation metrics from ICP.  

\begin{table}[htb]
    \small
    \centering
    \begin{tabular}{c|c|c|c|c|c}
    \hline
    L & Id & Team's Vol. & GT Vol. & Ch. w/ t.m & Ch. w/o t.m  \\
    \hline
\multirow{8}{*}{E}   & 1 & 40.06 & 38.53 & 1.63 & 85.40\\
                        & 2 & 216.9 & 280.36 & 7.12 & 111.47\\
                        & 3 & 278.86 & 249.67 & 13.69 & 172.88\\
                        & 4 & 279.02 & 295.13 & 2.03 & 61.30\\
                        & 5 & 395.76 & 392.58 & 13.67 & 102.14\\
                        & 6 & 205.17 & 218.44 & 6.68 & 150.78\\
                        & 7 & 372.93 & 368.77 & 4.70 & 66.91\\
                        & 8 & 186.62 & 173.13 & 2.98 & 152.34\\
\hline
                        
\multirow{7}{*}{M} &9 & 224.08 & 232.74 & 3.91 & 160.07\\
                        & 10 & 153.76 & 163.09 & 2.67 & 138.45\\
                        & 11 & 80.4 & 85.18 & 3.37 & 151.14\\
                        & 13 & 363.99 & 308.28 & 5.18 & 147.53\\
                        & 14 & 535.44 & 589.83 & 4.31 & 89.66\\
\hline
\multirow{7}{*}{H}& 16& 163.13 & 262.15 & 18.06 & 28.33\\
                        & 17 & 224.08 & 181.36 & 9.44 & 28.94\\
                        & 18 & 25.4 & 20.58 & 4.28 & 12.84\\
                        & 19 & 110.05 & 108.35 & 11.34 & 23.98\\
                        & 20 & 130.96 & 119.83 & 15.59 & 31.05\\
    \hline
    \end{tabular}
    \caption{Quantitative comparison of the team's approach with ground truth using challenge dataset. The team evaluates their approach using Chamfer distance in $\times 10^{-3}$ with and without transformation metrics. Volumes in the table is in $cm^3$.}
    \label{tab:results_1}
\end{table}

For overall method performance, Table~\ref{tab:results_2} shows the MAPE and Chamfer distance with and without transformation metrics. 

\begin{table}[htb]
    \small
    \centering
    \begin{tabular}{c|cc|cc}
    \hline
    MAPE ↓ (\%) & \multicolumn{2}{c|}{Ch. w/ t.m ↓} & \multicolumn{2}{c}{Ch. w/o t.m ↓} \\
    & sum & mean & sum & mean \\
    \hline
    \textbf{10.973} & 0.130 & \textbf{0.007} & 1.715 & \textbf{0.095} \\
    \hline
    \end{tabular}
    \caption{Quantitative comparison of the team's approach with ground truth using challenge dataset. The team evaluates their approach using Chamfer distance with and without transformation metrics. The results show the mean and sum of the 18 scenes.}
    \label{tab:results_2}
\end{table}

Additionally, Fig.~\ref{fig:qualitative_results} shows the qualitative results on the one and few-shot 3D reconstruction from the challenge dataset. The figures show that the team's model excels in texture details, artifact correction, missing data handling, and color adjustment across different scene parts.

\paragraph{Limitations.}
Despite the promising results demonstrated by the team's method, several limitations need to be addressed in future work:
\begin{itemize}
    \item \textbf{Manual processes:} The current pipeline includes manual steps, such as providing a segmentation prompt and identifying scaling factors. These steps should be automated to enhance efficiency and reduce human intervention. This limitation arises from the necessity of using a reference object to compensate for missing data sources, such as Inertial Measurement Unit (IMU) data.
    \item \textbf{Input requirements:} The team's method requires extensive input information, including food masks and depth data. Streamlining the necessary inputs would simplify the process and potentially increase its applicability in varied settings.
    \item \textbf{Complex backgrounds and objects:} The team has not tested their method in environments with complex backgrounds or on highly intricate food objects. Applying their approach to datasets with more complex food items, such as the Nutrition5k \cite{thames2021nutrition5k} dataset, would be challenging and could help identify corner cases that need to be addressed.
    \item \textbf{Capturing complexities:} The method has not been evaluated under different capturing complexities, such as varying distances between the camera and the food object, different camera speeds, and other scenarios as defined in the Fruits and Vegetables \cite{steinbrener2023learning} dataset. These factors could significantly impact the performance and robustness of the team's method.
    \item \textbf{Pipeline complexity:} For one-shot neural rendering, the team currently uses the One-2-3-45 \cite{liu2024one} method. However, they aim to use only the 2D diffusion model, Zero123 \cite{liu2023zero}, in their pipeline to reduce complexity and improve the efficiency of their approach.
\end{itemize}

%% file: sec/ININ-VIAUN.tex
\section{Second Place Team - ININ-VIAUN}
\label{sec: ININ-VIAUN}

\begin{figure*}[h]
    \centering
    \includegraphics[width=\textwidth]{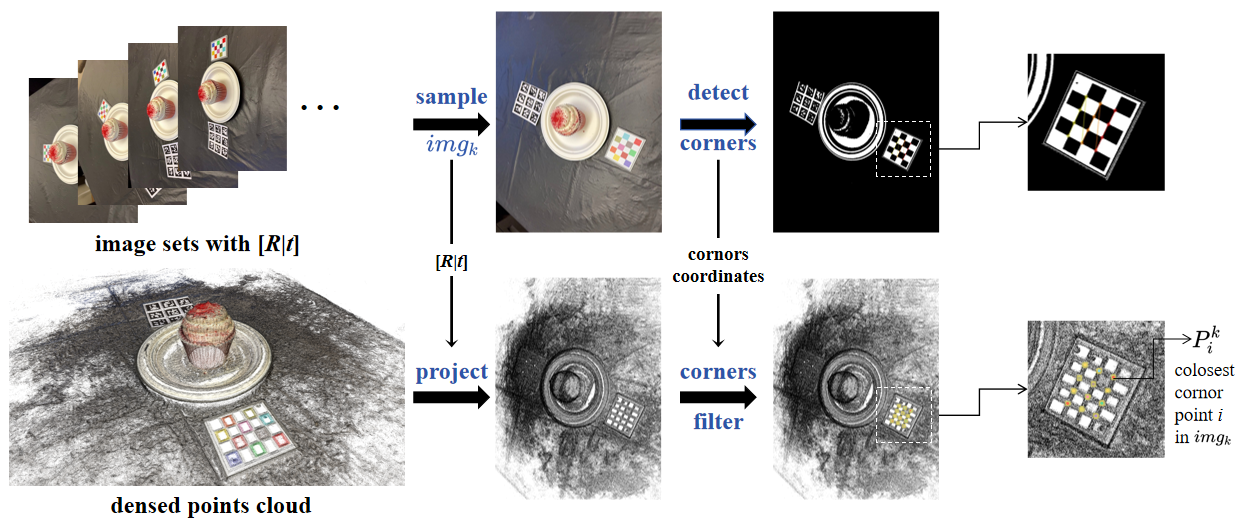}
    \caption{The pipeline of scale factor estimation.}
    \label{pipeline_scale}
\end{figure*}

\subsection{Methodology}

In this section, this team provides a detailed explanation of their proposed network, demonstrating how to progress from the original images to the final mesh models step by step. The code is available at \url{https://github.com/BITyia/cvpr-metafood}.

\subsubsection{Scale factor estimation}
The pipeline for coordinate-level scale factor estimation is shown in Figure \ref{pipeline_scale}. The team follows a corner projection matching method. Specifically, using the COLMAP\cite{schonberger2016structure} dense model, the team obtains the pose of each image as well as dense point cloud information. For any image \(\text{img}_k\) and its extrinsic parameters \([R/t]_k\), the team first performs a threshold-based corner detection with the threshold set to 240. This allows them to obtain the pixel coordinates of all detected corners. Subsequently, using the intrinsic parameters \(k\) and the extrinsic parameters \([R/t]_k\), the point cloud is projected onto the image plane. Based on the pixel coordinates of the corners, the team can identify the closest point coordinates \(P_i^k\) for each corner, where \(i\) represents the index of the corner. Thus, they can calculate the distance between any two corners as follows:

\begin{equation}
D_{ij}^k = \sqrt{(P_i^k - P_j^k)^2} \quad \forall i \neq j
\label{eq:distance}
\end{equation}

To determine the final computed length of each checkerboard square in image \(k\), the team takes the minimum value of each row of the matrix \(D^k\) (excluding the diagonal) to form the vector \(d^k\). The median of this vector is then used. The final scale calculation formula is given by Equation \ref{eq:scale}, where 0.012 represents the known length of each square (1.2 cm):

\begin{equation}
\text{scale} = \frac{0.012}{\frac{1}{n} \sum_{i=1}^n \text{med}(d^k)}
\label{eq:scale}
\end{equation}

\subsubsection{3D Reconstruction}
\begin{figure*}[h]
    \centering
    \includegraphics[width=\textwidth]{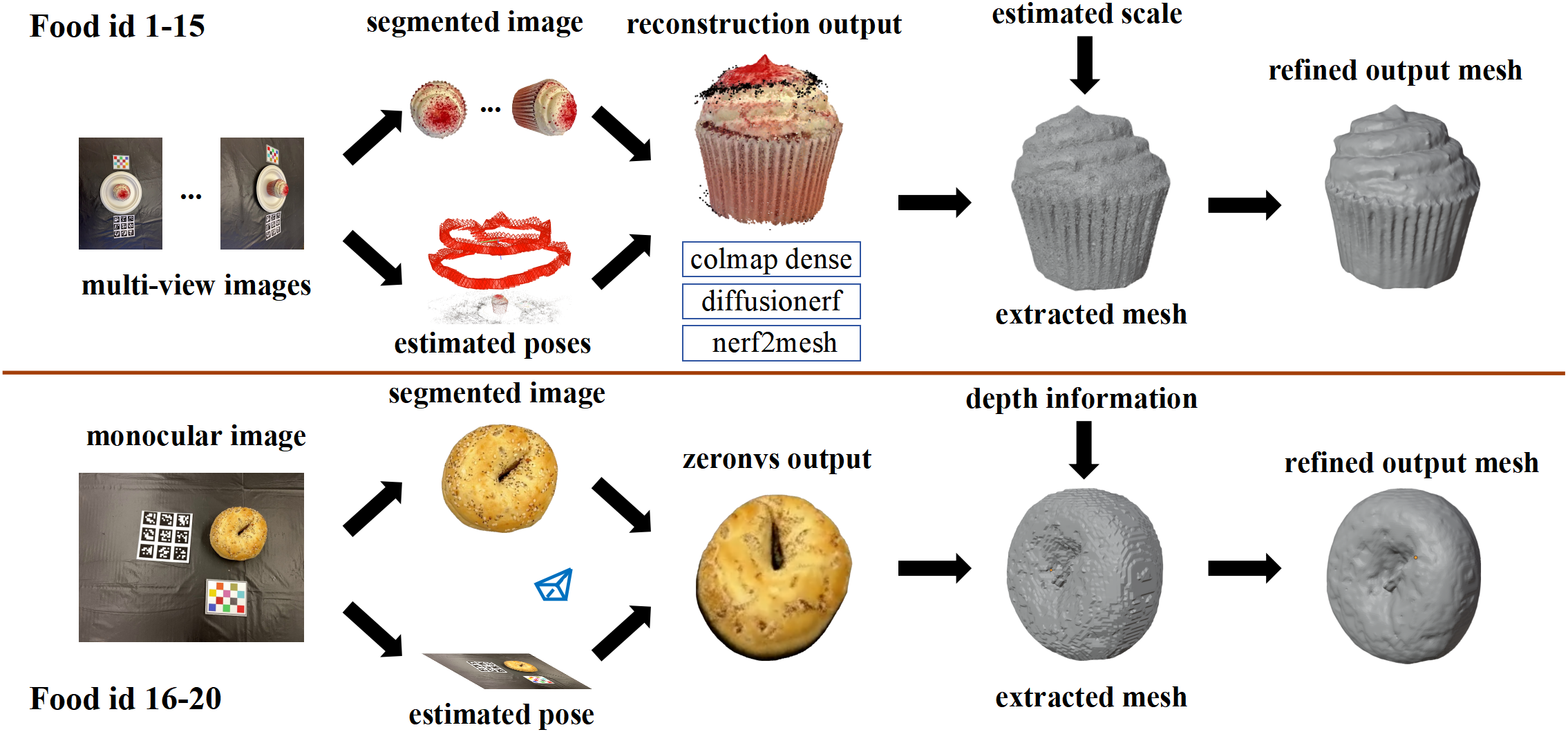} 
    \caption{The pipeline of 3d reconstruction.}
    \label{pipeline_recon}
\end{figure*}

The pipeline for 3D reconstruction is shown in Figure \ref{pipeline_recon}. Considering the differences in input viewpoints, the team utilizes two pipelines to process the first fifteen objects and the last five single view objects.

For the first fifteen objects, the team uses COLMAP\cite{schonberger2016structure} to estimate the poses and segment the food using the provided segment masks in the dataset. Then, they apply advanced multi-view 3D reconstruction methods to reconstruct the segmented food. In practice, the team employs three different reconstruction methods: COLMAP\cite{schonberger2016structure}, DiffusioNeRF\cite{wynn2023diffusionerf}, and NeRF2Mesh\cite{tang2023delicate}. They select the best reconstruction results from these methods and extract the mesh from the reconstructed model. Next, they scale the extracted mesh using the estimated scale factor. Finally, they apply some optimization techniques to obtain a refined mesh. 

For the last five single-view objects, the team experiments with several single-view reconstruction methods, such as Zero123\cite{liu2023zero}, Zero123++\cite{shi2023zero123++}, One2345\cite{liu2024one}, ZeroNVS\cite{sargent2024zeronvs}, and DreamGaussian\cite{tang2023dreamgaussian}. They choose ZeroNVS\cite{sargent2024zeronvs} to obtain a 3D food model consistent with the distribution of the input image. In practice, they use the intrinsic camera parameters from the fifteenth object and employ an optimization method based on reprojection error to refine the extrinsic parameters of the single camera. However, due to the limitations of single-view reconstruction, the team needs to incorporate depth information from the dataset and the checkerboard in the monocular image to determine the size of the extracted mesh. Finally, they apply optimization techniques to obtain a refined mesh.

\subsubsection{Mesh refinement}
\begin{figure}[h]
    \centering
    \includegraphics[width=\linewidth]{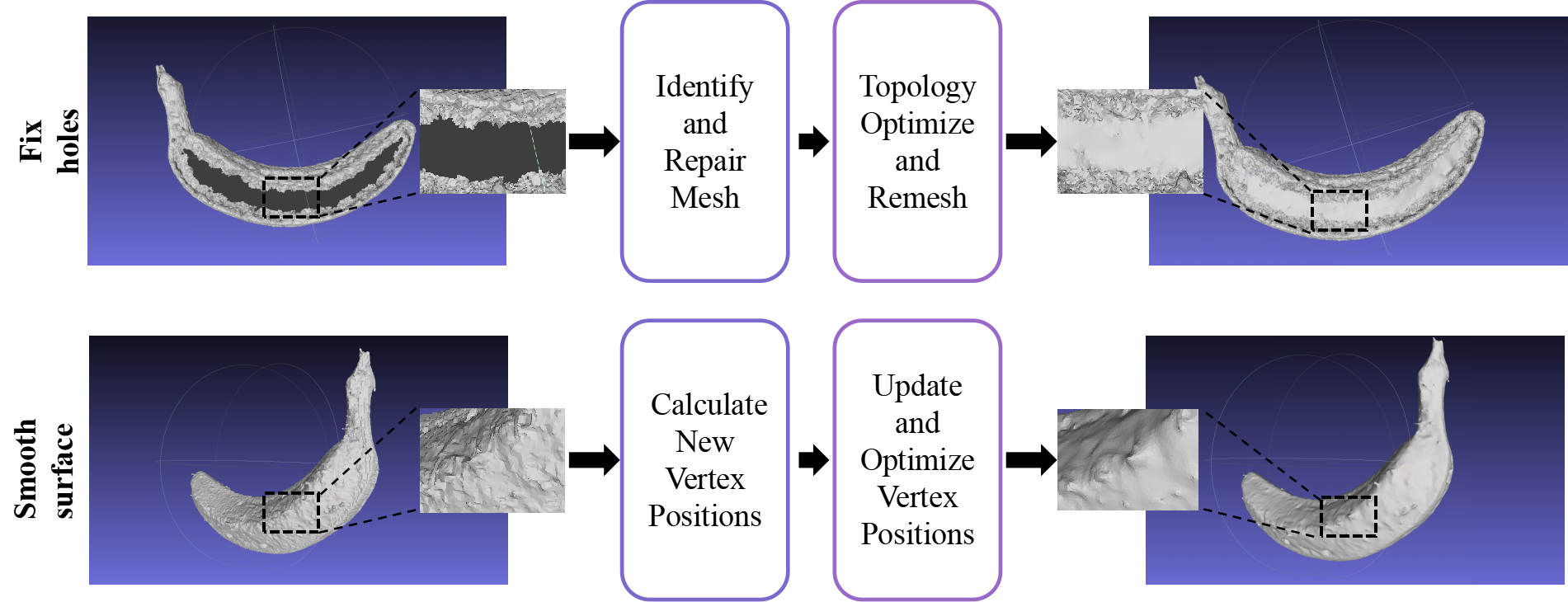} 
    \caption{Mesh refinement.}
    \label{mesh_refine}
\end{figure}

In the 3D Reconstruction phase, the team observes that the model's results often suffer from low quality due to the presence of holes on the object surface and substantial noise, as illustrated in Figure \ref{mesh_refine}.

To address the holes, the team employs MeshFix\cite{attene2010lightweight}, an optimization method based on computational geometry. For surface noise, they utilize Laplacian Smoothing\cite{locality_aware_smoothing} for mesh smoothing operations. The Laplacian Smoothing method works by adjusting the position of each vertex to the average of its neighboring vertices:

\[ 
v_i^{(new)} = v_i^{(old)} + \lambda \left( \frac{1}{N} \sum_{j \in \mathcal{N}(i)} v_j^{(old)} - v_i^{(old)} \right) 
\tag{3}
\]

In their implementation, the team sets the smoothing factor \( \lambda \)  to 0.2 and performs 10 iterations.

\subsection{Experimental Results}

\subsubsection{Estimated scale factor}
The scale factors estimated using the method described earlier are shown in Table \ref{tab_scale}. Each image and the corresponding reconstructed 3D model yield a scale factor, and the table presents the average scale factor for each object.

\begin{table}[h]
  \centering
  \begin{tabular}{@{}ccc@{}}
    \toprule
    Object Index & Food Item & Scale Factor \\
    \midrule
    1 & Strawberry & 0.060058 \\
    2 & Cinnamon bun & 0.081829 \\
    3 & Pork rib & 0.073861 \\
    4 & Corn & 0.083594 \\
    5 & French toast & 0.078632 \\
    6 & Sandwich & 0.088368 \\
    7 & Burger & 0.103124 \\
    8 & Cake & 0.068496 \\
    9 & Blueberry muffin & 0.059292 \\
    10 & Banana & 0.058236 \\
    11 & Salmon & 0.083821 \\
    13 & Burrito & 0.069663 \\
    14 & Hotdog & 0.073766 \\
    \bottomrule
  \end{tabular}
  \caption{Estimated scale factors.}
  \label{tab_scale}
\end{table}

\subsubsection{Reconstructed meshes}
\begin{figure}[h]
  \centering
   \includegraphics[width=\linewidth]{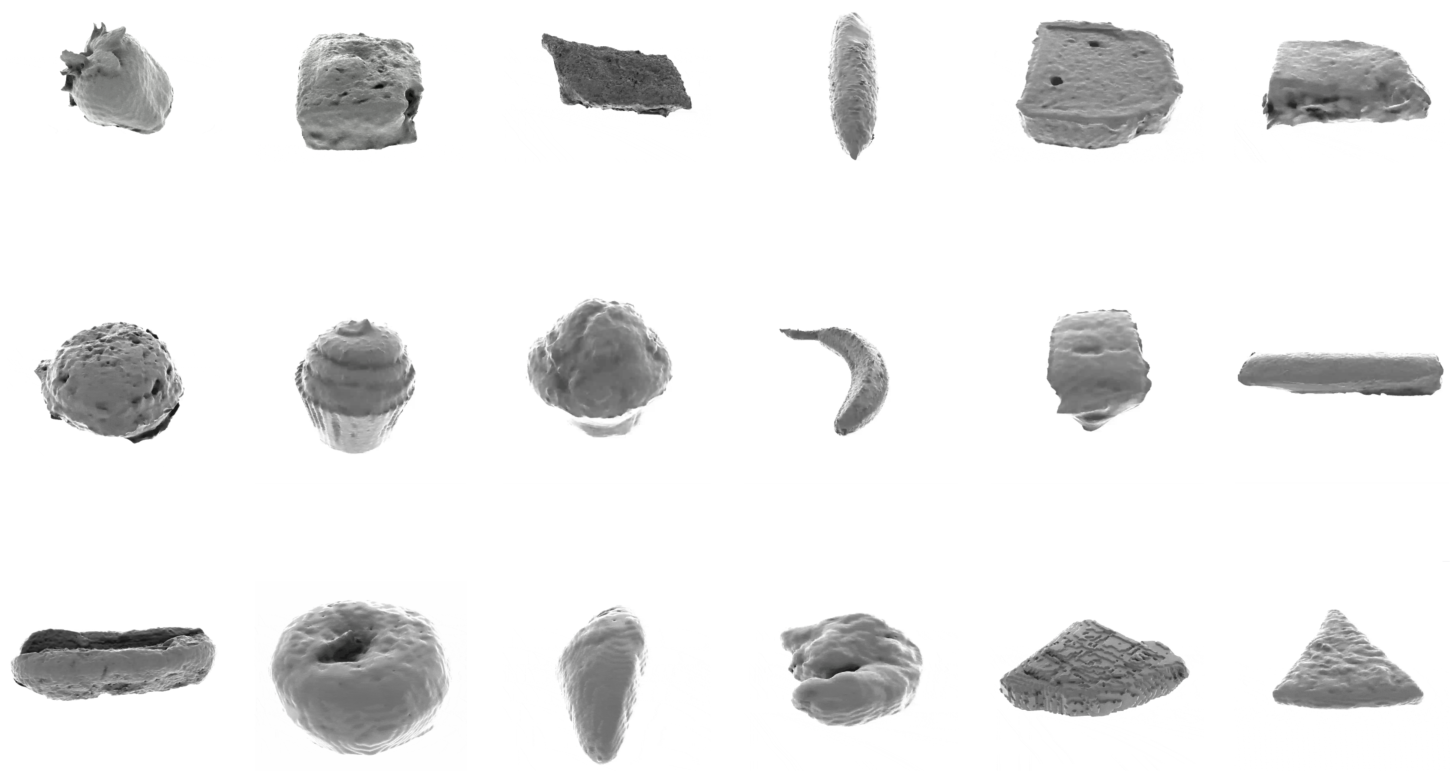}
   \caption{Refined Meshes.}
   \label{recon}
\end{figure}

The refined meshes obtained using the methods described earlier are shown in Figure \ref{recon}. The predicted model volumes, ground truth model volumes, and the percentage errors between them are shown in Table \ref{tab_volume}.

\begin{table}[h]
  \centering
  \setlength{\tabcolsep}{10pt}
  \begin{tabular}{@{}cccc@{}}
    \toprule
    Object & Predicted & Ground &  Error \\
    Index & volume & truth & percentage \\
    \midrule
    1 & 44.51 & 38.53 & 15.52 \\
    2 & 321.26 & 280.36 & 14.59 \\
    3 & 336.11 & 249.67 & 34.62\\
    4 & 347.54 & 295.13 & 17.76\\
    5 & 389.28 & 392.58 & 0.84\\
    6 & 197.82 & 218.44 & 9.44\\
    7 & 412.52 & 368.77 & 11.86\\
    8 & 181.21 & 173.13 & 4.67\\
    9 & 233.79 & 232.74 & 0.45\\
    10 & 160.06 & 163.09 & 1.86\\
    11 & 86.0 & 85.18 & 0.96\\
    13 & 334.7 & 308.28 & 8.57\\
    14 & 517.75 & 589.83 & 12.22\\
    16 & 176.24 & 262.15 & 32.77\\
    17 & 180.68 & 181.36 & 0.37\\
    18 & 13.58 & 20.58 & 34.01\\
    19 & 117.72 & 108.35 & 8.64\\
    20 & 117.43 & 119.83 & 20.03\\
    \bottomrule
  \end{tabular}
  \caption{Metric of volume. The unit is cubic millimeters.}
  \label{tab_volume}
\end{table}

\subsubsection{Alignment}
\begin{figure}[h]
  \centering
   \includegraphics[width=\linewidth]{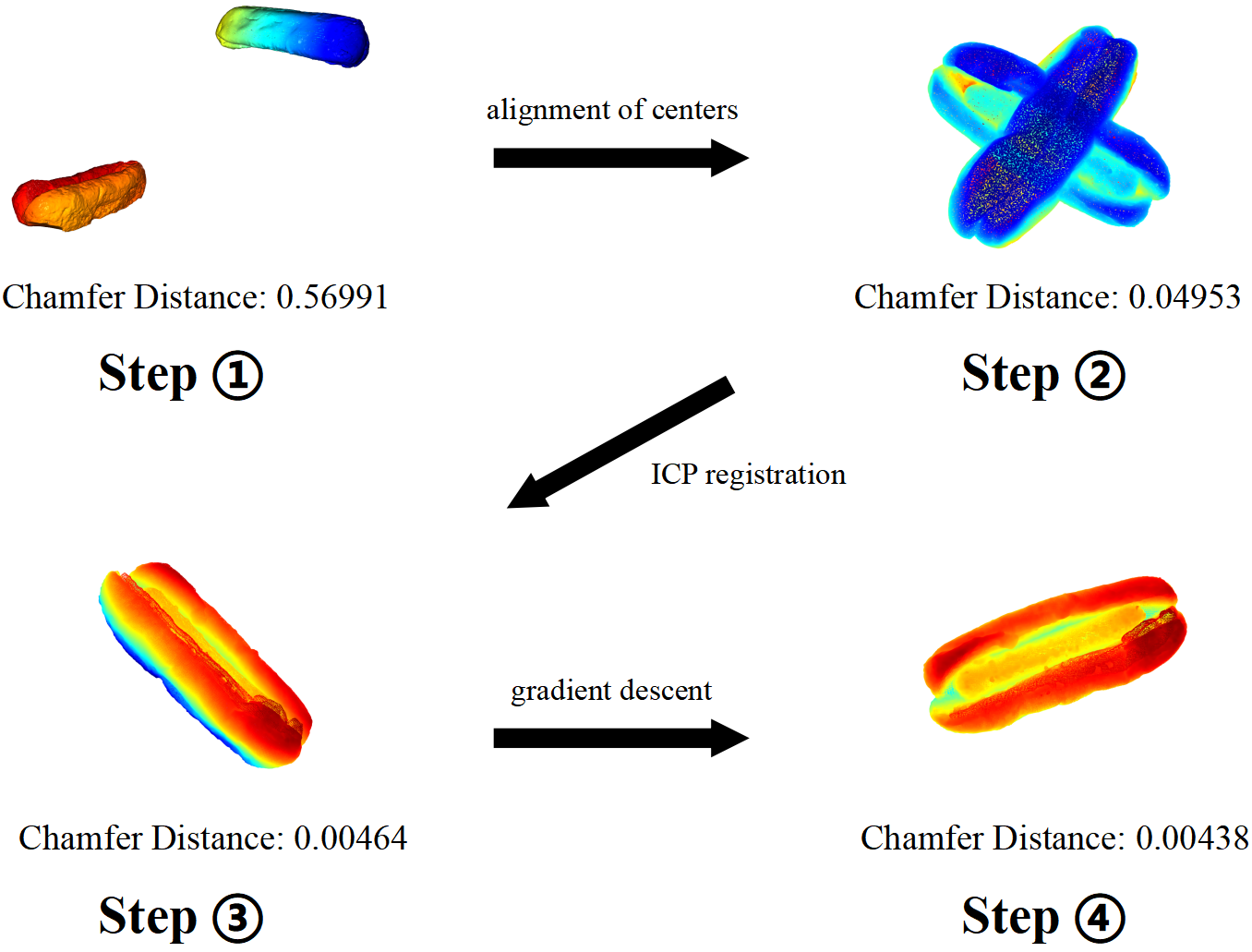}
   \caption{The process of aligning object 14.}
   \label{align}
\end{figure}

The team designs a multi-stage alignment method for evaluating reconstruction quality. Figure \ref{align} illustrates the alignment process for Object 14. First, the team calculates the central points of both the predicted model and the ground truth model, and moves the predicted model to align the central point of the ground truth model. Next, they perform ICP registration for further alignment, significantly reducing the Chamfer distance. Finally, they use gradient descent for additional fine-tuning, and obtain the final transformation matrix.

The total Chamfer distance between all 18 predicted models and the ground truths is 0.069441169.

\begin{figure*}[htbp]
    \centering
    \includegraphics[width=1. \linewidth]{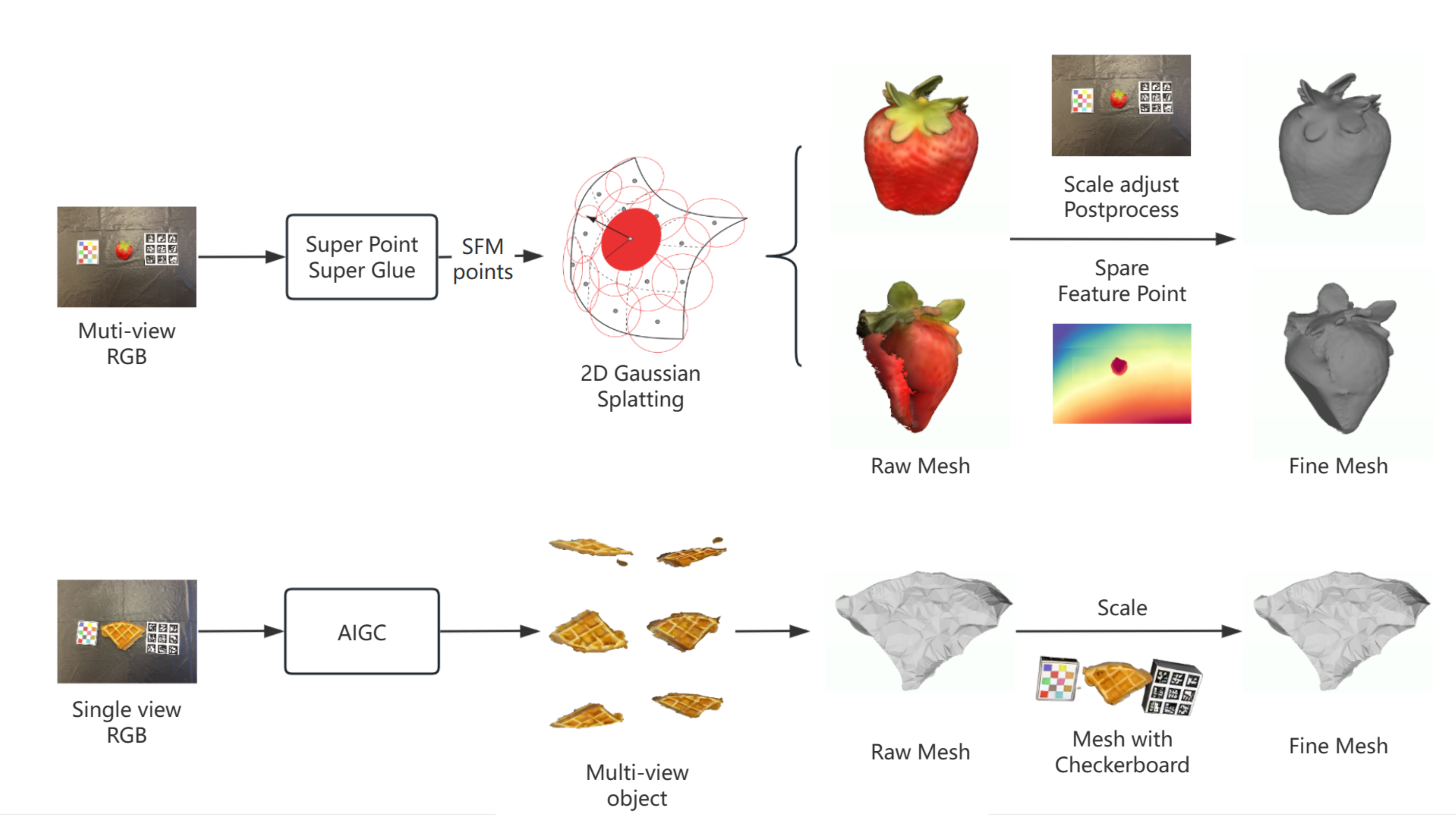}
    \caption{For multi-view image inputs, COLMAP integrated with SuperPoint and SuperGlue generates SFM points, which are used to create the initial Gaussian. Using 2D Gaussian Splatting, we obtain the mesh of the observed object. Subsequently, we adjust the mesh size and fit the unobserved underside of the object using the RGB checkerboard and depth maps. Finally, a complete and realistic food mesh is produced. For single-view input, we use the AIGC method to generate multi-view images consistent with the input image in 3D using a multi-view diffusion model. Then, using the Sparse-view Large Reconstruction Model, we directly predict the mesh.Lastly, using the simultaneously reconstructed checkboard, adjust the size of the food. }
    \label{fig:design}
\end{figure*}

%% file: sec/FoodRiddle.tex
\section{Best 3D Mesh Reconstruction Team - FoodRiddle}
\label{sec: FoodRiddle}
\begin{figure}[h]
    \centering
    \includegraphics[width=.85 \linewidth]{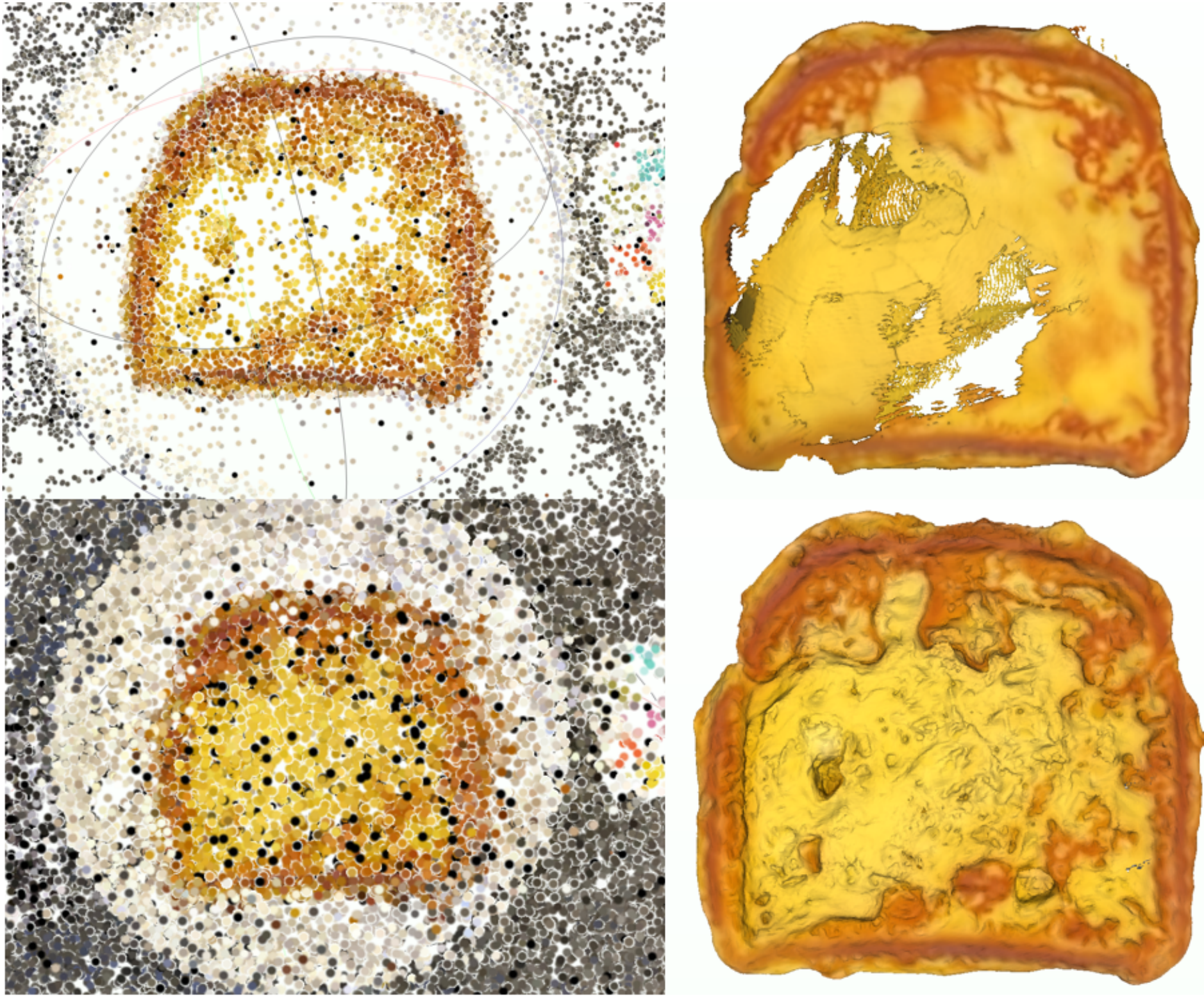}
    \caption{The top left image showcases the vanilla COLMAP sfm points. The light-colored areas on the bread have fewer feature points, leading to the incomplete mesh in the top-right image.However, by integrating SuperPoint and SuperGlue into COLMAP, more interest points are obtained, resulting in an excellent final mesh, as shown in the bottom-right image. }
    \label{fig:scenario}
\end{figure}

\subsection{Methodology}

To achieve high-quality food mesh reconstruction, the team designed two pipeline processes as shown in Figure~\ref{fig:design}. For simple and medium cases, they employed a structure-from-motion approach to determine the pose of each image, followed by mesh reconstruction. Subsequently, a series of post-processing steps were implemented to recalibrate scale and enhance mesh quality. For cases with only a single image, the team utilized image generation methods to aid in model generation.

\subsubsection{Multi-View Reconstruction}

For Structure from Motion (SfM), the team extended the state-of-the-art method COLMAP \cite{schoenberger2016sfm} by incorporating SuperPoint \cite{8575521} and SuperGlue \cite{sarlin2020superglue} methodologies. This significantly mitigated the issue of sparse keypoints in weakly textured scenes, as shown in Figure \ref{fig:scenario}.

For mesh reconstruction, the team's method is based on 2D Gaussian Splatting\cite{huang20242d}, which utilizes a differentiable 2D Gaussian renderer and incorporates regularization terms for depth distortion and normal consistency. The Truncated Signed Distance Function (TSDF) results are used to generate a dense point cloud.

In the post-processing stage, the team applied filtering and outlier removal techniques, identified the contour of the supporting surface, and projected the lower mesh vertices onto the supporting surface. They used the reconstructed checkerboard to rectify the scale of the model and used Poisson reconstruction to generate a watertight, complete mesh of the subject.

\subsubsection{Single-View Reconstruction}

For 3D reconstruction from a single image, the team employed state-of-the-art methods such as LGM\cite{tang2024lgm}, Instant Mesh\cite{xu2024instantmesh}, and One-2-3-45\cite{liu2023one2345} to generate an initial prior mesh. This prior mesh was then jointly corrected with depth structure information.

To adjust the scale, the team estimated the object's length using the checkerboard as a reference, assuming the object and the checkerboard are on the same plane. They then projected the 3D object back onto the original 2D image to recover a more accurate scale of the object.

\subsection{Experimental Results}

Through a process of nonlinear optimization, the team sought to identify a transformation that minimizes the Chamfer distance between their mesh and the ground truth mesh. This optimization aimed to align the two meshes as closely as possible in three-dimensional space. Upon completion of this process, the average Chamfer distance across the final reconstructions of the 20 objects amounted to 0.0032175 meters. As shown in Table \ref{table:errors}, Team FoodRiddle achieved the best scores for both multi-view and single-view reconstructions, outperforming other teams in the competition. The source code is available at \url{https://github.com/jlyw1017/FoodRiddle-MetaFood-CVPR2024}. 
\begin{table}[htbp]
\centering
\scalebox{0.95}{
\begin{tabular}{ccc}
\hline
{Team} & {Multi-view (1-14)} & {Single-view (16-20)} \\ \hline
\textbf{FoodRiddle} & \textbf{0.036362} & \textbf{0.019232} \\ 
{ININ-VIAUN} & 0.041552 & 0.027889 \\ 
{VolETA} & 0.071921 & 0.058726 \\
\hline
\end{tabular}
}
\caption{Total Errors for Different Teams on Multi-view and Single-view Data. Team FoodRiddle has the best score.}
\label{table:errors}
\end{table}

%% file: sec/5_conclusion.tex
\section{Conclusion}
\label{sec: conclusion}
In this report, we review and summarize the methods and results of the MetaFood CVPR Workshop challenge on 3D Food Reconstruction. The challenge aimed to advance 3D reconstruction techniques by focusing on food items, addressing the unique challenges posed by varying textures, reflective surfaces, and complex geometries typical in culinary subjects.

The competition utilized 20 diverse food items, captured under different conditions and with varying numbers of input images, specifically designed to challenge participants in developing robust reconstruction models. The evaluation was based on a two-phase process, assessing both portion size accuracy through Mean Absolute Percentage Error (MAPE) and shape accuracy using the Chamfer distance metric.

Out of all participating teams, three made it to the final submission, showcasing a range of innovative solutions. Team VolETA won the first place with the overall best performance on both Phase-I and Phase-II. Followed by team ININ-VIAUN who won the second place. In addition, FoodRiddle team demonstrated superior performance in Phase-II, indicating a competitive and high-caliber field of entries for 3D mesh reconstruction. The challenge has successfully pushed the boundaries of 3D food reconstruction, demonstrating the potential for accurate volume estimation and shape reconstruction in nutritional analysis and food presentation applications. The innovative approaches developed by the participating teams provide a solid foundation for future research in this field, potentially leading to more accurate and user-friendly methods for dietary assessment and monitoring.